\documentclass[10pt,twocolumn,letterpaper]{article}

\usepackage{iccv}
\usepackage{times}
\usepackage{epsfig}
\usepackage{graphicx}
\usepackage{amsmath}
\usepackage{amssymb}

\usepackage{booktabs}
\usepackage{multirow}

\usepackage{amsbsy}
\usepackage{array}

\usepackage{subfigure}

\usepackage{float}

\usepackage[pagebackref=true,breaklinks=true,letterpaper=true,colorlinks,bookmarks=false]{hyperref}

\iccvfinalcopy 

\def\iccvPaperID{****} 

\ificcvfinal\fi

\begin{document}

\title{Few-shot Fine-tuning is All You Need for Source-free Domain Adaptation}

\author{Suho Lee$^{1}$\thanks{Equal contribution}
\and
Seungwon Seo$^{1}$\footnotemark[1]
\and
Jihyo Kim$^{1}$
\and
Yejin Lee$^{2}$
\and
Sangheum Hwang$^{1}$\thanks{Corresponding author}
\and
\small $^1$ Department of Data Science, Seoul National University of Science and Technology, Republic of Korea\\
\small $^2$ NAVER Z Corporation, Republic of Korea\\
{\tt\small \{swlee,swseo,jihyo.kim\}@ds.seoultech.ac.kr} \hspace{3pt} {\tt\small yejinl@naverz-corp.com} \hspace{3pt} {\tt\small shwang@seoultech.ac.kr} 
}
\maketitle

\ificcvfinal\thispagestyle{empty}\fi

\begin{abstract}
   Recently, source-free unsupervised domain adaptation (SFUDA) has emerged as a more practical and feasible approach compared to unsupervised domain adaptation (UDA) which assumes that labeled source data are always accessible. However, significant limitations associated with SFUDA approaches are often overlooked, which limits their practicality in real-world applications. These limitations include a lack of principled ways to determine optimal hyperparameters and performance degradation when the unlabeled target data fail to meet certain requirements such as a closed-set and identical label distribution to the source data. All these limitations stem from the fact that SFUDA entirely relies on unlabeled target data. We empirically demonstrate the limitations of existing SFUDA methods in real-world scenarios including out-of-distribution and label distribution shifts in target data, and verify that none of these methods can be safely applied to real-world settings. Based on our experimental results, we claim that fine-tuning a source pretrained model with a few labeled data (e.g., 1- or 3-shot) is a practical and reliable solution to circumvent the limitations of SFUDA. Contrary to common belief, we find that carefully fine-tuned models do not suffer from overfitting even when trained with only a few labeled data, and also show little change in performance due to sampling bias. Our experimental results on various domain adaptation benchmarks demonstrate that the few-shot fine-tuning approach performs comparatively under the standard SFUDA settings, and outperforms comparison methods under realistic scenarios. Our code is available at \url{https://github.com/daintlab/fewshot-SFDA}.
\end{abstract}

\section{Introduction}
Deep neural networks (DNNs) have recently achieved remarkable performance in a broad range of applications. However, these successes typically rely on the assumption that the distributions of the training and test data are identical, which is often violated in practice. Deep learning models are known to suffer from significant performance degradation when the test data distribution deviates from the training data distribution~\cite{wang2016training, johnson2019survey}. This problem, known as domain shift~\cite{quinonero2008dataset, branco2016survey}, limits the general applicability of DNNs, and has motivated a variety of research topics aimed at mitigating the domain shift problem.

\begin{figure}[t]
    \centering
    \hspace{8mm}
    \includegraphics[width=0.4\textwidth]{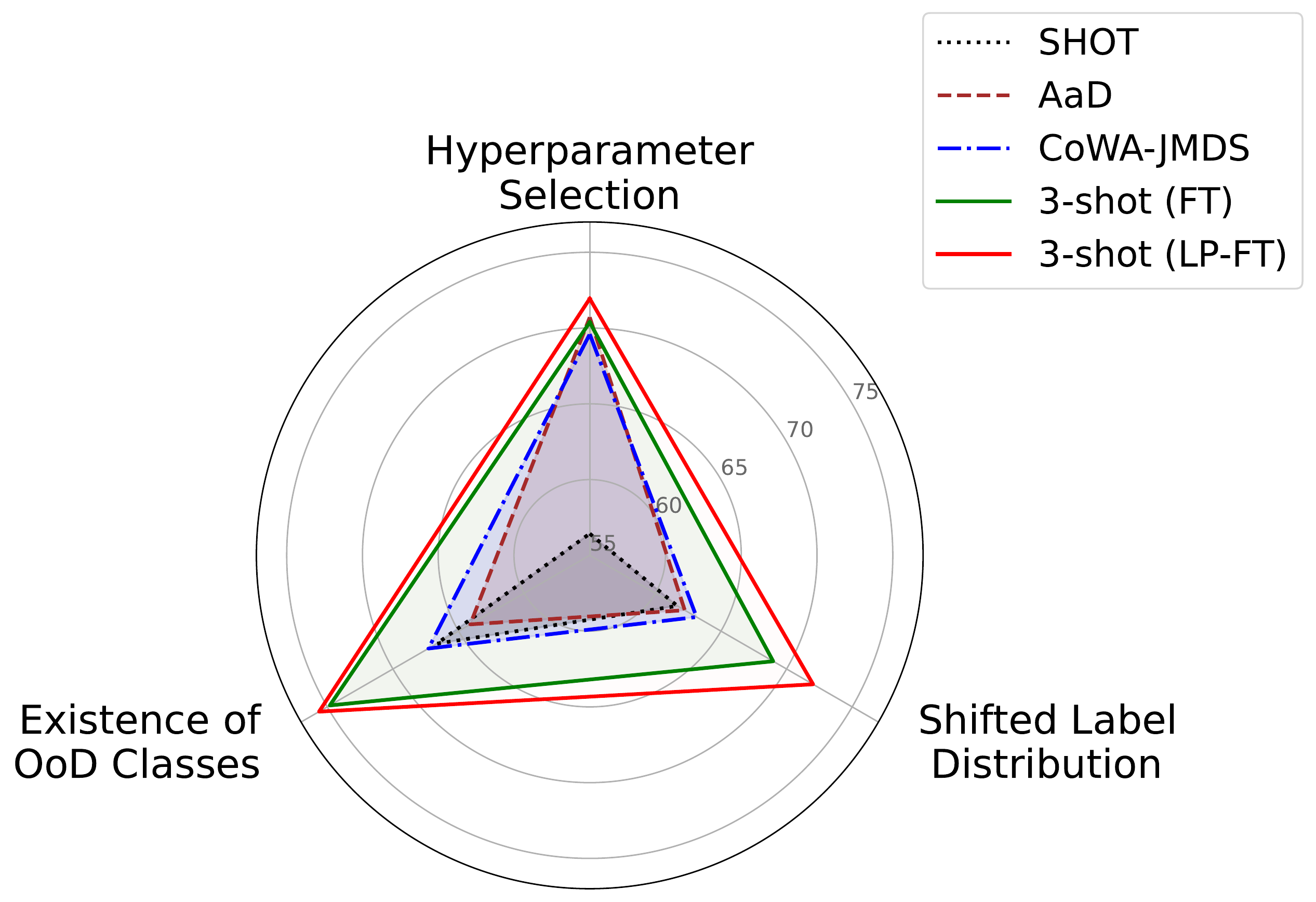}
    \caption{Overall comparison of SFUDA and few-shot SFDA methods under the issues that may be encountered in real-world scenarios. We use average target domain accuracy as a performance metric. In the ``Hyperparameter Selection'' experiments, we measure the performance of SFUDA methods using hyperparameters selected through Soft Neighborhood Density~\cite{saito2021tune}.}
    \label{fig:radar_chart}
\end{figure}

Unsupervised domain adaptation (UDA) represents a prominent research direction that addresses the issue of domain shift. Specifically, UDA aims to effectively transfer knowledge of a source domain to a target domain, relying on both labeled source data and unlabeled target data. By doing so, the performance of the model can be significantly enhanced as it allows for a better alignment of the learned representations with the target domain.
Over the past few years, UDA has gained considerable attention across various computer vision tasks, including but not limited to image classification~\cite{ganin2016domain, tzeng2017adversarial, wang2020self, li2021transferable}, semantic segmentation~\cite{wang2020classes,pandey2020unsupervised,yang2021exploring}, object detection~\cite{oza2021unsupervised, vs2021mega}, and person re-identification~\cite{mekhazni2020unsupervised, chen2020deep}. However, the underlying assumption of UDA that both the source and target data are simultaneously available during the adaptation process is often violated for various reasons, such as data privacy, storage and transmission costs, or the computational burden associated with training on a large-scale source data~\cite{fang2022source}.

To overcome these limitations of UDA, a new paradigm of domain adaptation has emerged in recent years, namely, source-free unsupervised domain adaptation (SFUDA)~\cite{liang2020we, liang2021source, lee2022confidence, yang2021generalized, yang2022attracting}. The primary difference between UDA and SFUDA lies in the accessibility to the source data. In SFUDA, the source data is not available during the adaptation phase. Instead, SFUDA fully leverages a source pretrained model to adapt to the target domain using unlabeled data. Due to the growing concerns regarding data privacy and intellectual property, as well as the recent trend towards large-scale pretraining, the source-free assumption has been considered practical and reasonable by the machine learning community. However, considering that SFUDA solely relies on the unlabeled target data that is given in advance, we would like to pose the following question: \textit{Do current SFUDA methods adequately address the realistic domain adaptation scenarios?}

Given that SFUDA entirely relies on the target unlabeled data, it may encounter certain challenges and limitations when utilized for real-world applications. First, it is important to note that there is currently no principled approach for selecting optimal hyperparameters for existing SFUDA methods. Properly validating a model with unlabeled data is problematic, making it difficult to determine which hyperparameters are best for a given target domain. Second, it cannot be guaranteed that the closed-set assumption will always hold in practice. In other words, the unlabeled data provided for domain adaptation may contain out-of-distribution (OoD) classes that are not utilized during source model training. As most current SFUDA methods have been developed under the closed-set assumption, they are prone to performance degradation in OoD scenarios. Finally, even if unlabeled training data satisfy the closed-set assumption, their label (class) distribution may differ significantly from that of the source training data. For example, a major class in the source data may become a minor class in the target data. Under such a label distribution shift, the knowledge of a source pretrained model may not be effectively leveraged during the adaptation process. 
Several studies have tackled those challenges in the context of UDA~\cite{tan2020class, shi2022pairwise} and SFUDA~\cite{li2021imbalanced, sun2022source}. Nevertheless, to the best of our knowledge, none of these studies can resolve all of the aforementioned issues, indicating that we still need a practical solution for source-free domain adaptation (SFDA).

In this work, we argue that fine-tuning a source pretrained model using a small amount of labeled data (e.g., 1- or 3-shot) from the target domain is a more practical and reliable alternative. Acquiring a few labeled data from the target domain is cost-effective and allows for the circumvention of the problematic scenarios associated with relying solely on unlabeled data. To support our claim, we first demonstrate that the performance of existing popular SFUDA methods is highly sensitive to hard-to-determine hyperparameters involved in training, and it significantly degrades when unlabeled target data have OoD classes or substantially different class distributions. Surprisingly, we observe that fine-tuning a source pretrained model with only a few labeled data can yield comparable or better performance without the risk of overfitting.
We also show that learning the decision boundary (i.e., the attached classifier) prior to fine-tuning the entire model, known as LP-FT~\cite{kumar2022fine} proposed for OoD generalization, can further improve performance. Notably, few-shot fine-tuning is not significantly affected by sampling bias even under the challenging settings of 1-shot training and 1-shot validation. This is in contrast to the results that ImageNet-1k~\cite{deng2009imagenet} pretrained models tend to severely overfit to a few labeled data. We hypothesize that this phenomenon arises from the high degree of semantic similarity between source and target domains in domain adaptation problems.

Figure~\ref{fig:radar_chart} summarizes the comparison results on the OfficeHome~\cite{venkateswara2017deep} dataset between state-of-the-art SFUDA methods, SHOT~\cite{liang2020we}, AaD~\cite{yang2022attracting} and CoWA-JMDS~\cite{lee2022confidence}, and LP-FT as well as end-to-end fine-tuning (denoted as FT) with three labeled data per class (i.e., $3$-shot fine-tuning), considering the aforementioned issues. We can observe that even $3$-shot fine-tuning outperforms state-of-the-art SFUDA methods, which indicates that few-shot fine-tuning can be a viable alternative to SFUDA methods.

Our main contributions are summarized as follows: 
\begin{itemize}
    \item We analyze the challenges that hinder the practical applicability of SFUDA in real-world scenarios, including the ambiguity in hyperparameter selection, the presence of OoD data in the target data, the discrepancy in class distributions between the source and target data, and show all recent methods are susceptible to these issues.
    \item  We demonstrate that fine-tuning with a few labeled target data, namely few-shot SFDA, can be a practical and effective solution that naturally avoids the aforementioned challenges associated with SFUDA.
    \item Our extensive experiments on five benchmarks show that the few-shot SFDA methods (i.e., FT and LP-FT) perform comparably to or surpass state-of-the-art SFUDA methods in various realistic scenarios using only a small amount of labeled target data.
\end{itemize}

\section{Related Work}
SFUDA methods can be broadly divided into three categories: generative approaches~\cite{li2020model, kurmi2021domain, ijcai2021p0402}, pseudo-labeling approaches~\cite{liang2020we, liang2021source, lee2022confidence}, and neighborhood clustering approaches~\cite{yang2021generalized, yang2021exploiting, yang2022attracting}. The generative approaches attempt to compensate for the absence of source data by synthesizing target-style images~\cite{li2020model}, proxy source data~\cite{kurmi2021domain}, or feature prototypes that represent the source data~\cite{ijcai2021p0402}. However, learning generative models can be challenging and may encounter issues such as mode collapse or a lack of diversity in the generated samples. 
Pseudo-labeling and neighborhood clustering approaches have also emerged as promising techniques for SFUDA, exhibiting effective and straightforward adaptation to the target domain.

SHOT~\cite{liang2020we}, a representative pseudo-labeling method, uses self-supervised pseudo-labeling and information maximization strategies to align target features to the source feature space while freezing the source classifier. Subsequently, SHOT$++$~\cite{liang2021source} has further enhanced the performance by combining SHOT with a self-supervised auxiliary task and MixMatch~\cite{berthelot2019mixmatch}, a semi-supervised learning method, to fully exploit pseudo labels. In addition, CoWA-JMDS~\cite{lee2022confidence} has introduced a new confidence score, Joint Model-Data Structure (JMDS), that utilizes both source and target domain knowledge to differentiate sample-wise importance for pseudo labels.
On the other hand, several methods based on neighborhood clustering have been proposed in the context of SFUDA. G-SFDA~\cite{yang2021generalized} and NRC~\cite{yang2021exploiting} basically consist of a neighborhood clustering term for prediction coherence and a marginal entropy term for prediction diversity. AaD~\cite{yang2022attracting} encourages consistent predictions across neighbors, while dispersing the predictions of potentially dissimilar features.

Although SFUDA has shown remarkable progress, it faces several challenges in real-world scenarios. One of the primary challenges is related to hyperparameter selection. Recently, Saito \etal~\cite{saito2021tune} have tackled this problem by proposing an unsupervised validation criterion called Soft Neighborhood Density (SND), which serves as a proxy for target domain performance, and demonstrated its effectiveness on UDA methods. SND measures the density of soft neighborhoods by calculating the entropy of the similarity distribution between data points. However, it remains unclear whether this unsupervised validation method performs well for existing SFUDA methods.
Another potential challenge for SFUDA methods is when the target domain includes OoD data, such as data belonging to classes that are not of interest. This scenario is known as open-set UDA, and several studies have proposed SFUDA methods to mitigate performance degradation in this case~\cite{liang2021umad,saito2020universal,kundu2020universal,yang2022one}. Even in closed-set settings where the label spaces of the source and target domains are consistent, the label distributions may differ, which presents another challenge for SFUDA methods. Some recent studies have addressed this issue, and demonstrated the effectiveness of the proposed algorithms in the presence of intra-domain class imbalance and inter-domain label shift~\cite{li2021imbalanced,sun2022source}.
However, although all of these issues may inherently limit the applicability of SFUDA methods, previous studies have mainly focused on each challenge individually and developed advanced algorithms to address a specific problem.

Recently, test-time adaptation (TTA)~\cite{nado2020evaluating, wang2020tent, iwasawa2021test} has been studied, where the adaptation to the target domain is performed during inference. TTA is often considered to be more practical than SFUDA, as it does not require pre-collected target domain data. However, TTA faces similar challenges as SFUDA, since it also relies on unlabeled target data for adaptation. For example, Niu~\etal~\cite{niu2023towards} showed that the performance of TTA methods is highly dependent on the size and label distribution of the test batch data given for the adaptation. 

Similar to our work, the feasibility of few-shot SFDA has been examined in Zhang~\etal~\cite{zhang2022few}. However, their focus is mainly on comparing with TTA methods, and they lack a thorough analysis of the practicality of SFUDA. Additionally, the performance of the compared baselines, including FT, may have been underestimated due to the absence of a proper validation process (see Section~\ref{sec:compare_tl}).

\section{Pitfalls Overlooked in SFUDA Methods} \label{sec:three}
Unlike traditional UDA, SFUDA assumes that source data are not available during the adaptation process. Denote $\mathcal{X}$ as the input space and $\mathcal{Y}$ as the label space. In this work, we consider a $K$-way image classification task, i.e., $\mathcal{Y}=\{1,...,K\}$. Let $\mathcal{D}_s$ and $\mathcal{D}_t$ be the source and target domains, respectively, with $\mathcal{D}_s \neq \mathcal{D}_t$ due to domain shift. Given a source model $f_s{\,:\,}\mathcal{X}_s \xrightarrow{} \mathcal{Y}_s$ pretrained on $\{x_i^s,y_i^s\}_{i=1}^{N_s}$ from $\mathcal{D}_s$ and $N_t$ unlabeled data $\{x_i^t\}_{i=1}^{N_t}$ from $\mathcal{D}_t$, the goal of SFUDA is to obtain a model $f_t{\,:\,}\mathcal{X}_t \xrightarrow{} \mathcal{Y}_t$ that perform well on $\mathcal{D}_t$.
In this section, we raise concerns about the practical applicability of SFUDA, relying exclusively on unlabeled target data. If label information is absent from the target data, practitioners may encounter a range of challenges when utilizing SFUDA methods in real-world applications. These challenges include, but are not limited to, selecting optimal hyperparameters, dealing with OoD scenarios, and addressing label distribution shifts. 

To identify potential shortcomings of the SFUDA approach, we conduct a thorough investigation into the performance of state-of-the-art SFUDA methods, SHOT~\cite{liang2020we}, CoWA-JMDS~\cite{lee2022confidence}, and AaD~\cite{yang2022attracting}, under the aforementioned scenarios. We experiment on four popular benchmarks for domain adaptation: Office31~\cite{saenko2010adapting}, OfficeHome~\cite{venkateswara2017deep}, VLCS~\cite{fang2013unbiased}, and TerraIncognita (Terra)~\cite{beery2018recognition}. For all experiments, we use $80\%$ of the target domain data as training data and the remaining $20\%$ as test data. We evaluate the model on each source-target domain pair and report the average performance across all pairs. For example, in the Office31 dataset which includes three domains, Amazon (A), DSLR (D), and Webcam (W), we evaluate performance for all six source$\rightarrow$target tasks: \{A$\rightarrow$D, A$\rightarrow$W, D$\rightarrow$A, D$\rightarrow$W, W$\rightarrow$A, W$\rightarrow$D\}. All the results are averaged over three random runs (refer to Section~\ref{sec:exp-set} for the detailed dataset description and experimental settings). Although these SFUDA methods have demonstrated promising performance on certain benchmarks, our experimental results highlight the potential limitations of relying solely on unlabeled data for SFDA in real-world settings.

\subsection{Ambiguity in Hyperparameter Selection}\label{sec:hp}
Hyperparameters are typically selected by evaluating target performance on validation data, and choosing the values that give the best performance. However, since SFUDA only has access to unlabeled target data, it is challenging to estimate generalization performance during validation in a principled manner when a classification task is given. Nevertheless, most SFUDA methods lack a clear explanation of how hyperparameters are chosen, and the impact of hyperparameter selection is often overlooked. Recently, an unsupervised validation criterion, SND~\cite{saito2021tune}, has been proposed for UDA as a proxy for performance on the target domain. However, it remains unclear whether SND can be effectively applied to a variety of SFUDA methods.

We conduct experiments to show that it is challenging to decide proper hyperparameter values for SFUDA in two aspects. Firstly, we investigate the sensitivity of SFUDA methods to hyperparameters. To this end, we compare the performance of current SFUDA methods while varying a learning rate used during target adaptation, which is a non-trivial hyperparameter to determine in practice. For each method, we test five different learning rates, including $lr \times k, k \in \{0.1, 0.5, 1, 5, 10\}$, where $lr$ is the learning rate reported in the literature. Secondly, we explore the possibility of finding the optimal learning rate in an unsupervised manner, using the SND criterion. We compute the SND of each configuration on the target test data to evaluate its effectiveness in hyperparameter selection for SFUDA.

\begin{figure}
    \centering
    \subfigure[Office31]{
        \centering
        \includegraphics[width=0.46\columnwidth]{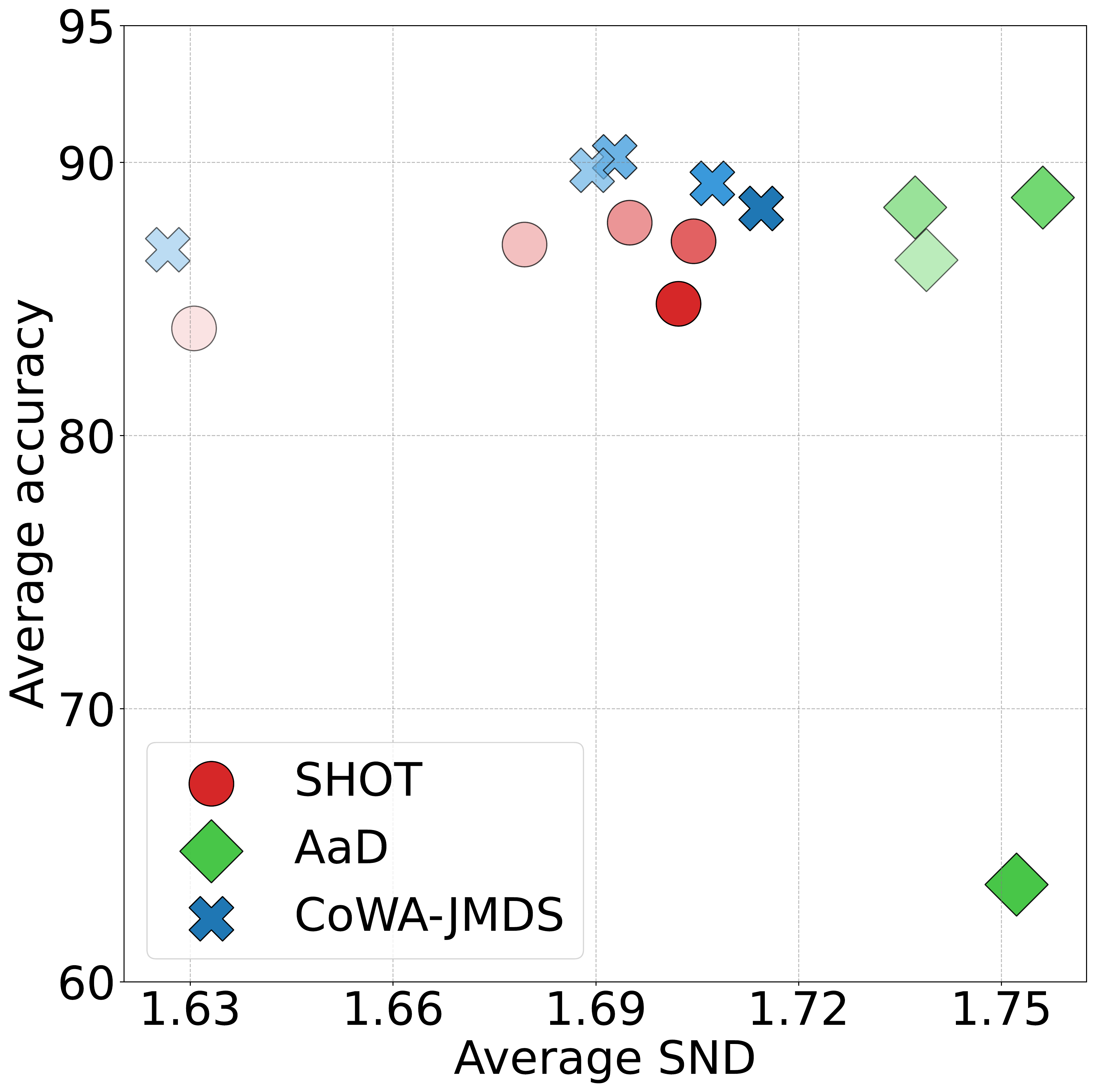}
        \label{fig:snd_office31}
        }
    \subfigure[OfficeHome]{
        \centering
        \includegraphics[height=0.46\columnwidth]{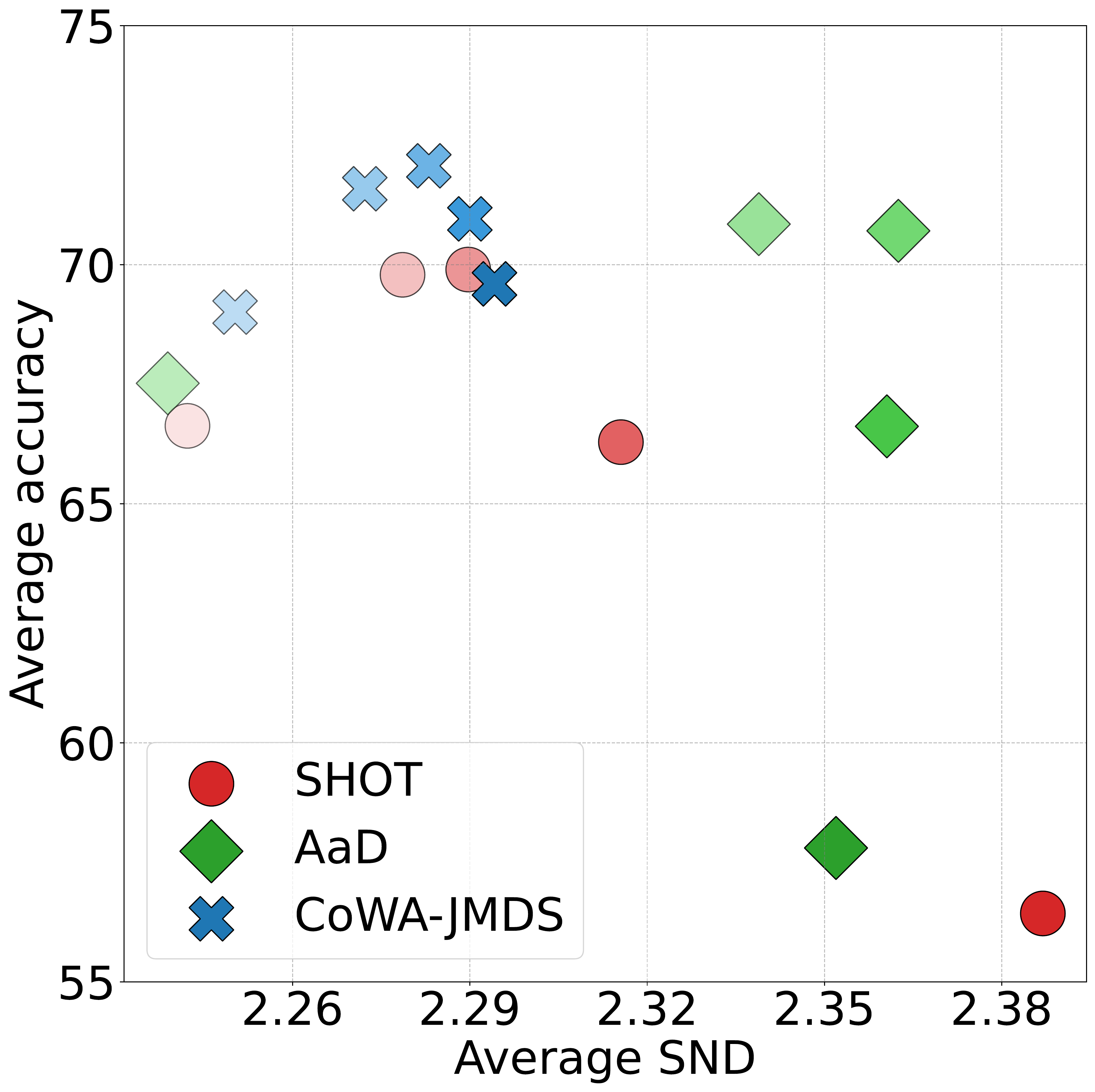}
        \label{fig:snd_officehome}
        }
    \caption{Average accuracy and SND values for SFUDA methods on Office31 and OfficeHome datasets. The brightness of each point indicates the learning rate magnitude, where brighter colors correspond to lower learning rates.}
    \label{fig:snd}
\end{figure}

Figure~\ref{fig:snd} shows the average accuracy and average SND values of each model for Office31\footnote {The result of AaD with $lr\times 10$ is omitted for clear visualization. The average accuracy and SND are $38.09$ and $2.14$, respectively.} and OfficeHome. Each point represents a model trained by a specific SFUDA method with a particular learning rate. We observe that the performance of each SFUDA method significantly varies depending on the learning rate. For example, if the worst learning rate is selected for SHOT, the average accuracy drops by up to $13.47\%$ on OfficeHome compared to the best performance. Similarly, the performance of AaD drops by up to $50.62\%$ on Office31 compared to the best performance. This shows that choosing an appropriate learning rate is crucial for the SFUDA methods. Also, SND does not show any positive correlation to the target performance on both datasets. Especially, AaD and SHOT models show the lowest accuracy on Office 31 and OfficeHome, respectively, when the learning rate is chosen based on SND. This indicates that the current unsupervised validation criterion is inadequate and there is currently no definitive method for selecting hyperparameters in SFUDA.

\begin{table}[t]
    \begin{center}
    \resizebox{0.48\textwidth}{!}{%
    \begin{tabular}{p{75pt}ccccccc}
        \toprule
        \multirow{2}{*}{\textbf{Method}} & \multicolumn{2}{c}{\textbf{Config} $\textbf{1}$} & \multicolumn{2}{c}{\textbf{Config} $\textbf{2}$} & \multicolumn{2}{c}{\textbf{Config} $\textbf{3}$}\\
        & \textit{Clean} & \textit{OoD} & \textit{Clean} & \textit{OoD} & \textit{Clean} & \textit{OoD} \\
        \midrule
        No adapt & \multicolumn{2}{c}{$64.41$} & \multicolumn{2}{c}{$59.12$} & \multicolumn{2}{c}{$72.01$}\\
        SHOT~\cite{liang2020we} & $73.18$ & $66.66$ & $68.88$ & $64.20$ & $82.57$ & $77.57$\\
        CoWA-JMDS~\cite{lee2022confidence} & $75.13$ & $67.30$ & $71.62$ & $66.19$ & $83.99$ & $79.41$\\
        AaD~\cite{yang2022attracting} & $70.34$ & $64.11$ & $67.36$ & $60.85$ & $80.82$ & $76.81$ \\
        \bottomrule
    \end{tabular}
    }
    \end{center}
    \caption{Average accuracy of SFUDA methods in \textit{Clean} and \textit{OoD} settings. ``No adapt" refers to the direct evaluation of the source pretrained model without any target domain training.}
    \label{tab:ood}
\end{table}

\subsection{Performance Degrades in OoD Scenario}
When collecting unlabeled data from the target domain in practice, some of the data may belong to unknown classes (i.e., OoD data). The inclusion of OoD data can introduce unexpected bias to the model and can potentially degrade performance. Additionally, effectively rejecting OoD data in the target domain using only the source pretrained model poses a significant challenge, further complicating the task of SFUDA. Although there are several SFUDA methods that have addressed this scenario, called open-set DA, their performance may not be satisfactory as reported in~\cite{fang2022source}. Additionally, these methods often require modifications to the algorithmic process based on whether OoD data exist in the unlabeled target data~\cite{liang2020we,lee2022confidence}. For instance, SHOT needs an additional procedure to exclude uncertain data from the loss computation, particularly for the OoD scenario. CoWA-JMDS achieves optimal performance under the OoD scenario by using different training configurations than those for the vanilla closed-set setting. However, it is infeasible to determine whether the target data contains OoD classes or not. Moreover, it is unclear if these specific modifications also work well in the closed-set setting. Therefore, modifying the algorithm procedure or training configuration may not be practical in real-world applications.

Here, we focus on the intrinsic performance drop of SFUDA methods in the OoD scenario. To this end, we conduct experiments on the OfficeHome dataset~\cite{venkateswara2017deep} with and without OoD data in the unlabeled target training data. Firstly, we randomly select $25$ classes as known classes and treat the remaining classes as unknowns. Next, we train a model on the source domain data consisting only of known classes. We then adapt the source pretrained model to the target domain using each SFUDA method under two conditions: (1) \textit{Clean setting} where the target training data only contain known classes and (2) \textit{OoD setting} where the target training data contain both known and unknown classes. Finally, we evaluate each method on the target test data consisting of known classes. All hyperparameters are set to the same values as those reported in the corresponding papers.

In all three configurations where known classes are randomly chosen, all SFUDA methods show performance degradation when OoD data are included during training as presented in Table~\ref{tab:ood}. For instance, CoWA-JMDS exhibits a decrease in performance of $7.83\%$ in Configuration 1. These results indicate that existing SFUDA methods are not robust against OoD scenarios, and the presence of OoD data can have a negative effect on their performance. This weakness of SFUDA methods makes it challenging for practitioners to use them in real-world settings, where OoD data can emerge unexpectedly.

\begin{table}[t]
    \begin{center}
    \resizebox{0.35\textwidth}{!}{%
    \begin{tabular}{p{75pt}cccccc|c}
        \toprule
        \textbf{Method} & \textbf{Office31} & \textbf{VLCS} & \textbf{Terra} \\
        \midrule
        No adapt & $77.63$ & $56.74$ & $29.18$ \\
        SHOT~\cite{liang2020we} & $87.79$ & $65.58$ & $\textbf{25.72}$ \\
        CoWA-JMDS~\cite{lee2022confidence} & $90.20$ & $72.31$ & $\textbf{28.60}$ \\
        AaD~\cite{yang2022attracting} & $88.71$ & $\textbf{56.35}$ & $\textbf{22.62}$ \\
        \midrule
        Bhatt. distance & $0.02$ & $0.10$ & $0.31$ \\
        \bottomrule
    \end{tabular}
    }
    \end{center}
    \caption{Average performance of SFUDA methods on benchmarks with different levels of label distribution shifts. Large Bhattacharyya distance indicates a strong shift. \textbf{Bold} represents lower performance than ``No adapt".}
    \label{tab:imbal}
\end{table}

\subsection{Negative Effects under Label Distribution Shift}\label{sec:pitfall_label}
Even if all unlabeled target training data correspond to classes in the source domain, there is no guarantee that their class distribution will be balanced or similar to that of the source data. For instance, in the context of developing object recognition models for autonomous driving, datasets collected from different countries may contain more images of bicycles or pedestrians than the source domain, causing a label distribution shift. Several studies have attempted to address this label distribution shift issue and proposed advanced algorithms for UDA~\cite{tan2020class} or SFUDA~\cite{li2021imbalanced}. Although these methods can alleviate this problem to some extent, they still suffer from performance degradation as the degree of label shift becomes more severe~\cite{li2021imbalanced}.

We evaluate the performance degradation of recent SFUDA methods on three benchmarks, Office31, VLCS, and Terra, which have different degrees of label distribution shifts across their domains. To quantify the degrees of the shifts, we calculate the Bhattacharyya distance of empirical label distributions obtained from each domain and compute their average across all source-target domain pairs. Specifically, we measure the degree of label distribution shift between domain $s$ and $t$ using the following equation:
\begin{equation}
    D_{(s,t)} = -\ln\left(\sum_{y\in \mathcal{Y}}\sqrt{P_s(y)P_t(y)}\right)~,
\end{equation}
where $P_s(y)$ and $P_t(y)$ represent the empirical label distributions of domain $s$ and $t$, respectively.
The hyperparameters for Office31 are configured based on the existing literature. For VLCS and Terra which are not considered in previous studies, we utilize one of the hyperparameter configurations for Office31 and OfficeHome from the literature, showing superior performance since a definitive approach for hyperparameter selection remains elusive, as discussed in Section~\ref{sec:hp}.
The average accuracy is reported for Office31, which has a relatively balanced label distribution. For VLCS and Terra containing substantial label distribution shifts, the per-class average accuracy is used instead, following~\cite{tan2020class,li2021imbalanced}.

Table~\ref{tab:imbal} demonstrates that all SFUDA methods show a negative effect on datasets with substantial label distribution shifts. Consistent with the results of previous works, SFUDA methods perform well on Office31, which is relatively balanced in terms of class distribution. However, for VLCS and Terra, some SFUDA methods have failed to enhance the performance of the source pretrained model. It is important to note that ``No adapt'' refers to the performance of the source model (i.e., not trained on the target domain). Especially for Terra, which has the most severe label distribution shift, none of the SFUDA methods prove effective, implying the possibility of negative effects of SFUDA methods in such scenarios.
Due to the risk of such negative outcomes, it is challenging to utilize these approaches in practice, given that we cannot anticipate the label distribution of the target domain beforehand.

\section{Few-shot Source-Free Domain Adaptation}

In Section~\ref{sec:three}, we present the challenges that arise due to the SFUDA assumption in three real-world scenarios. We now claim that fine-tuning the source pretrained model with a small amount of labeled target data can be a viable alternative for source-free domain adaptation. To demonstrate our claim, we compare few-shot fine-tuning methods with recent SFUDA methods using various domain adaptation benchmarks. 

Since the source and target domains often share considerable semantic information, preserving source pretrained knowledge during fine-tuning can be an effective strategy for achieving few-shot SFDA. In a recent study, it is demonstrated that na\"ive fine-tuning can distort the pretrained feature in a domain generalization setting. To mitigate this issue, a simple approach called LP-FT is proposed~\cite{kumar2022fine}. Specifically, LP-FT trains only a classification head and then performs end-to-end fine-tuning to alleviate the distortion of the pretrained feature. Thus, we also incorporate LP-FT in addition to na\"ive fine-tuning in our experiments.

\subsection{Experimental Settings} \label{sec:exp-set}

\paragraph{Datasets.} We utilize five benchmark datasets that are commonly used in domain adaptation research: Office31, OfficeHome, VisDA-C 2017, VLCS, and TerraIncognita.
\textbf{Office31}~\cite{saenko2010adapting} consists of objects commonly encountered in office settings. This dataset comprises three domains (Amazon, DSLR, and Webcam) containing $4,652$ images and $31$ categories.
\textbf{OfficeHome}~\cite{venkateswara2017deep} consists of common objects. This dataset contains four domains (Art, Clipart, Product, and Real World) containing $15,500$ images and $65$ categories.
\textbf{VisDA-C 2017 (VisDA)}~\cite{peng2017visda} is a large-scale dataset for synthetic-to-real domain adaptation across $12$ categories. This dataset contains $152k$ synthetic images for the source domain and $55k$ real-world images for the target domain.
\textbf{VLCS} dataset~\cite{fang2013unbiased} consists of common objects. This dataset comprises subsets from four different datasets (PASCAL VOC2007~\cite{everingham2009pascal}, LabelMe~\cite{russell2008label}, Caltech-101~\cite{li2004learning}, and SUN09~\cite{choi2010exploiting}) containing $10,729$ images and $5$ categories.
\textbf{TerraIncognita (Terra)}~\cite{beery2018recognition} consists of photographs of wild animals captured at different locations. The dataset comprises four domains (L100, L38, L43, and L46), containing $24,788$ images and $10$ categories.\footnote{We do not include cat and bird categories because they do not have enough images to conduct experiments with more than 5-shot.} For a fair comparison of FT, LP-FT, and SFUDA methods, we split the target domain into $8{:}2$ to construct the training and test dataset.\footnote{The standard practice for existing SFUDA methods involves assessing performance on the entire target domain data, which is also employed for training. For a rigorous comparison, we split the target data. Results obtained without the train/test split can be found in the supplementary material.} We consider a single source$\rightarrow$single target domain adaptation, and report the average performance over all source-target domain pairs. All experimental results are the average value obtained from three random runs.

\paragraph{Baselines.} We compare few-shot fine-tuning methods, FT and LP-FT, with six recent SFUDA methods: SHOT~\cite{liang2020we}, AaD~\cite{yang2022attracting}, G-SFDA~\cite{yang2021generalized}, NRC~\cite{yang2021exploiting}, CoWA-JMDS~\cite{lee2022confidence}, and SHOT$++$~\cite{liang2021source}. All SFUDA methods are reproduced using their officially released codes. For datasets not considered in the corresponding paper, such as VLCS and Terra, we select hyperparameters using the same approach as described in \ref{sec:pitfall_label}. 
To ensure a fair comparison, we adopt the same source pretrained model for all methods following \cite{liang2020we} except G-SFDA since it has a distinct source training procedure. For G-SFDA, we utilize the source pretrained model trained using the official codes.

\subsection{Implementation Details}

\paragraph{Model architecture.} We use the same network architecture utilized in~\cite{liang2020we}, which is also employed in the previous works we consider. Specifically, a feature extractor is followed by the bottleneck module consisting of fully connected layers, batch normalization~\cite{ioffe2015batch}, and weight normalization~\cite{salimans2016weight}. Consistent with prior studies~\cite{liang2020we,yang2022attracting,yang2021generalized,yang2021exploiting,lee2022confidence,liang2021source}, we adopt ResNet-101\cite{he2016deep} as the feature extractor for VisDA and ResNet-50 for other datasets.

\paragraph{Few-shot SFDA.} For both FT and LP-FT, we use the Adam~\cite{kingma2015adam} optimizer with sharpness-aware minimization~\cite{foret2021sharpnessaware} and set the batch size to $32$. The source pretrained model is fine-tuned for $1,000$ iterations. In the case of LP-FT, the classification head is trained for $1,000$ iterations before the end-to-end FT, resulting in a total of $2,000$ iterations. We employ simple data augmentation schemes as~\cite{gulrajani2021in}, including random resized cropping, horizontal flipping, color jittering, and gray scaling.

Given the limited labeled data available for few-shot SFDA, it is not feasible to collect a large amount of labeled data to be used for hyperparameter selection. Hence, we employ $1$-shot validation (i.e., a single labeled data per class) and select the hyperparameter that yields the lowest validation loss. Specifically, we search for the learning rate for few-shot SFDA methods in 3-shot configuration, exploring a set of values \{$1e$-$06$, $1e$-$05$, $1e$-$04$, $1e$-$03$\}. We apply the searched learning rate to all other few-shot configurations. Additionally, validation data and few-shot training data are randomly selected for each random run.

\begin{table}[t!]
    \begin{center}
    \resizebox{0.48\textwidth}{!}{%
    \begin{tabular}{lc|ccccc}
        \toprule
        \multicolumn{2}{l|}{\textbf{Method}} & \textbf{Office31} & \textbf{OfficeHome} & \textbf{VisDA} & \textbf{VLCS} & \textbf{Terra} \\
        \midrule
        \multicolumn{2}{l|}{No adapt} & $77.63$ ($0.68$) & $59.26$ ($0.57$) & $47.78$ ($0.88$) & $56.74$ ($0.94$) & $29.18$ ($0.08$) \\
        \multicolumn{2}{l|}{SHOT~\cite{liang2020we}} & $87.79$ ($1.45$) & $69.90$ ($0.43$) & $81.77$ ($0.58$) & $65.58$ ($2.53$) & $25.72$ $(1.05)$ \\
        \multicolumn{2}{l|}{SHOT$++$~\cite{liang2021source}} & $88.91$ ($0.75$) & $70.57$ ($0.50$) & \underline{$86.84$ ($0.31$)} & $64.27$ ($1.11$) & $27.07$ ($1.32$) \\
        \multicolumn{2}{l|}{AaD~\cite{yang2022attracting}} & $88.71$ ($0.77$) & $70.71$ ($0.96$) & $86.68$ ($0.47$) & $56.35$ ($1.11$) & $22.62$ ($2.75$) \\
        \multicolumn{2}{l|}{CoWA-JMDS~\cite{lee2022confidence}} & \underline{$90.20$ ($0.26$)} & \underline{$72.07$ ($0.74$)} & $86.06$ ($0.29$) & \underline{$71.30$ ($0.47$)} & $28.60$ ($3.09$) \\
        \multicolumn{2}{l|}{NRC~\cite{yang2021exploiting}} & $88.59$ ($1.10$) & $70.25$ ($0.20$) & $85.31$ ($0.63$) & $63.35$ ($3.96$) & $26.35$ ($2.53$) \\
        \multicolumn{2}{l|}{G-SFDA~\cite{yang2021generalized}} & $86.99$ ($0.78$) & $69.22$ ($0.47$) & $84.72$ ($0.21$) & $65.16$ ($0.13$) & \underline{$31.01$ ($2.15$)} \\
        \midrule
        \multirow{4}{*}{FT} & $1$ ($10$) & $82.88$ ($0.91$) & $64.45$ ($0.94$) & $81.20$ ($0.87$) & $64.39$ ($3.43$) & $\textcolor{red}{\textbf{39.17}}$ $\textcolor{red}{\textbf{(}\textbf{2.55}\textbf{)}}$ \\ %
        & $3$ ($20$) & $87.55$ ($0.61$) & $70.38$ ($0.26$) & $83.31$ ($1.19$) & $\textbf{71.53}$ \textbf{(}$\textbf{1.63}$\textbf{)} & $\textcolor{red}{\textbf{48.44}}$ $\textcolor{red}{\textbf{(}\textbf{1.68}\textbf{)}}$ \\
        & $5$ ($30$) & $\textbf{90.07}$ \textbf{(}$\textbf{0.46}\textbf{)}$ & $\textcolor{red}{\textbf{73.37}}$ $\textcolor{red}{\textbf{(}\textbf{0.19}\textbf{)}}$ & $84.28$ ($0.81$) & $\textbf{72.11}$ \textbf{(}$\textbf{0.89}$\textbf{)} & $\textcolor{red}{\textbf{54.61}}$ \textcolor{red}{\textbf{(}$\textbf{2.88}$\textbf{)}} \\
        & $10$ ($50$) & N/A & \textcolor{red}{$\textbf{76.99}$} \textcolor{red}{\textbf{(}$\textbf{0.41}$\textbf{)}} & $\textbf{85.93}$ \textbf{(}$\textbf{0.96}$\textbf{)} & $\textcolor{red}{\textbf{74.46}}$ \textcolor{red}{\textbf{(}$\textbf{1.17}$\textbf{)}} & N/A \\
        \midrule
        \multirow{4}{*}{LP-FT} & $1$ ($10$) & $82.85$ ($2.46$) & $63.63$ ($1.78$) & $81.24$ ($1.06$) & $\textbf{64.82}$ \textbf{(}$\textbf{1.50}\textbf{)}$ & $\textcolor{red}{\textbf{40.96}}$ \textcolor{red}{\textbf{(}$\textbf{2.82}$\textbf{)}} \\
        & $3$ $(20)$ & $\textbf{88.93}$ \textbf{(}$\textbf{0.69}$\textbf{)} & $\textbf{71.94}$ \textbf{(}$\textbf{1.10}$\textbf{)} & $83.53$ \textbf{(}$0.94$\textbf{)} & $\textcolor{red}{\textbf{73.04}}$ \textcolor{red}{\textbf{(}$\textbf{1.49}$\textbf{)}} & $\textcolor{red}{\textbf{48.98}}$ \textcolor{red}{\textbf{(}$\textbf{1.80}$\textbf{)}} \\
        & $5$ ($30$) & $\textcolor{red}{\textbf{90.93}}$ \textcolor{red}{\textbf{(}$\textbf{0.15}$\textbf{)}} & $\textcolor{red}{\textbf{74.55}}$ \textcolor{red}{\textbf{(}$\textbf{1.54}$\textbf{)}} & $84.82$ ($0.63$) & $\textcolor{red}{\textbf{73.07}}$ \textcolor{red}{\textbf{(}$\textbf{1.29}$\textbf{)}} & $\textcolor{red}{\textbf{55.40}}$ \textcolor{red}{\textbf{(}$\textbf{2.78}$\textbf{)}} \\
        & $10$ ($50$) & N/A & $\textcolor{red}{\textbf{78.58}}$ \textcolor{red}{\textbf{(}$\textbf{1.24}$\textbf{)}} & $\textbf{86.17}$ \textbf{(}$\textbf{0.63}$\textbf{)} & $\textcolor{red}{\textbf{74.52}}$ \textcolor{red}{\textbf{(}$\textbf{1.43}$\textbf{)}} & N/A \\
        \bottomrule 
    \end{tabular} 
    }
    \end{center}
    \caption{Average performance and standard deviation of SFUDA and few-shot fine-tuning methods over three random runs. The numbers next to FT and LP-FT represent the number of labeled images per class used for training. \underline{Underline} indicates the best performance among SFUDA methods. \textbf{Bold} and \textcolor{red}{\textbf{Bold Red}} indicate fine-tuning performance surpassing the average performance and the best performance of SFUDA methods, respectively. $10$-shot is not considered for Office31 and Terra because at least one of the classes has less than $10$ images.}
    \label{tab:main}
\end{table}

\subsection{Comparison with SFUDA and Few-shot SFDA} \label{sec:4.3}

\paragraph{Results under the vanilla SFDA settings.} We begin by comparing the existing SFUDA methods, FT and LP-FT under the standard SFDA settings, using the benchmarks without any modification. We report the per-class average accuracy for VisDA, VLCS, and Terra following~\cite{tan2020class,li2021imbalanced} and the average accuracy for the remaining datasets. For FT and LP-FT, we utilize $1$, $3$, $5$, and $10$ target images per class across all benchmarks, except for VisDA, which is a large-scale dataset. For VisDA, we use $10$, $20$, $30$, and $50$ target images per class, which is equivalent to $0.54\%$, $0.81\%$, $1.08\%$, and $1.35\%$, respectively, of the training set used for SFUDA methods.

As shown in Table~\ref{tab:main}, FT achieves competitive performance to SFUDA methods while utilizing only a few labeled data. Additionally, FT outperforms all of the SFUDA methods by a large margin with only a single data per class on Terra. LP-FT further enhances performance in most cases, which implies that preserving source pretrained features can be an effective strategy for few-shot SFDA. However, for VisDA, since the labeled data used for FT and LP-FT are significantly smaller than those used for SFUDA, $50$-shot (approximately $1.08\%$ of the entire training data) is required to achieve comparable performance. Notably, even $1$-shot FT substantially enhances the performance of the source pretrained model (i.e., ``No adapt'') on all datasets. This finding contradicts the common belief that training with only a few data often results in severe overfitting. We also observe that the performance variations (i.e., the standard deviations in Table~\ref{tab:main}) of FT and LP-FT are comparable to those of SFUDA methods, indicating that the sampling bias in few-shot SFDA is negligible. Detailed results for each dataset are available in the supplementary material.

\paragraph{Results under real-world scenarios.} Next, we compare FT and LP-FT with SFUDA methods under OoD data and label distribution shift settings. For the OoD data scenario, we conduct similar experiments as done in Section~\ref{sec:pitfall_label}. We randomly select $15$, $25$, and $6$ classes as known classes for Office31, OfficeHome, and VisDA, respectively. For the label distribution shift setting, we use the OfficeHome (RSUT)~\cite{tan2020class} and VisDA (RSUT) datasets~\cite{li2021imbalanced}, which are the imbalanced versions of OfficeHome and VisDA, respectively. The imbalanced datasets are created by subsampling the original datasets. As a consequence, the label distribution of the source domain is long-tailed and is the reversed version of that of the target domain~\cite{tan2020class}. We report the average accuracy for Office31 and OfficeHome in the OoD scenario, and the per-class average accuracy for the rest of the datasets. For the FT and LP-FT, we consider $1$- and $3$-shot training across all datasets except for VisDA and VisDA (RSUT), where we consider $3$- and $10$-shot training, respectively.

The results are shown in Table~\ref{tab:pitfalls}. As expected, we observe that FT and LP-FT with a few labeled data achieve better performance in most of the cases. When compared to SFUDA methods trained with OoD data, even $1$-shot FT outperforms all SFUDA methods on Office31. Also, $3$-shot FT can achieve better performance than the average performance of SFUDA methods on VisDA. In the presence of label distribution shift, the performance of all SFUDA methods degrades significantly compared to the results in Table~\ref{tab:main}, which is consistent with the observations in Section~\ref{sec:pitfall_label}. Furthermore, $3$-shot LP-FT surpasses the best-performing SFUDA method with a large margin on both VisDA (RSUT) and OfficeHome (RSUT). Overall, FT and LP-FT achieve superior performance than SFUDA methods in practical scenarios, which highlights the effectiveness of few-shot SFDA.

\begin{table}[t!]
    \begin{center}
    \resizebox{0.48\textwidth}{!}{%
    \begin{tabular}{ll|ccc|cc}
        \toprule
        & & \multicolumn{3}{c|}{\textbf{OoD}} & \multicolumn{2}{c}{\textbf{Imbalance}} \\
        \multicolumn{2}{l|}{\textbf{Method}} & \textbf{Office31} & \textbf{OfficeHome} & \textbf{VisDA} & \begin{tabular}[c]{@{}c@{}} \textbf{OfficeHome} \\ \textbf{(RSUT)} \end{tabular} & \begin{tabular}[c]{@{}c@{}} \textbf{VisDA} \\ \textbf{(RSUT)} \end{tabular} \\
        \midrule
        \multicolumn{2}{l|}{SHOT~\cite{liang2020we}} & $87.87$ ($0.98$) & $66.66$ ($0.50$) & $83.83$ ($0.91$) & $61.67$ ($0.81$) & $68.52$ ($2.98$) \\
        \multicolumn{2}{l|}{SHOT$++$~\cite{liang2021source}} & $88.65$ ($0.85$) & $66.31$ ($0.29$) & $83.62$ ($1.39$) & $59.49$ ($0.81$) & $65.49$ ($6.36$) \\
        \multicolumn{2}{l|}{AaD~\cite{yang2022attracting}} & $88.67$ ($1.09$) & $64.11$ ($1.28$) & \underline{$88.16$ ($1.50$)} & $62.24$ ($2.16$) & \underline{$69.82$ ($0.50$)} \\
        \multicolumn{2}{l|}{CoWA-JMDS~\cite{lee2022confidence}} & $89.45$ ($2.68$) & $67.30$ ($0.67$) & $79.45$ ($0.35$) & \underline{$63.14$ ($0.47$)} & $61.03$ ($2.89$) \\
        \multicolumn{2}{l|}{NRC~\cite{yang2021exploiting}} & $88.23$ ($2.29$) & \underline{$68.48$ ($0.89$)} & $85.86$ ($2.19$) & $57.98$ ($2.11$) & $69.20$ ($4.14$) \\
        \multicolumn{2}{l|}{G-SFDA~\cite{yang2021generalized}} & \underline{$89.97$ ($2.36$)} & $66.68$ ($1.06$) & $77.34$ ($0.21$) & $57.05$ ($1.70$) & $45.28$ ($3.32$) \\
        \midrule
        \multirow{2}{*}{FT} & $1$ ($3$) & $\textcolor{red}{\textbf{90.22}}$ \textcolor{red}{\textbf{(}$\textbf{2.27}$\textbf{)}} & $\textbf{67.45}$ \textbf{(}$\textbf{0.87}$\textbf{)} & $\textbf{84.31}$ ($\textbf{0.29}$) & $\textbf{61.63}$ \textbf{(}$\textbf{2.40}$\textbf{)} & $\textcolor{red}{\textbf{73.52}}$ \textcolor{red}{\textbf{(}$\textbf{3.98}$\textbf{)}} \\
        & $3$ ($10$) & $\textcolor{red}{\textbf{93.25}}$ \textcolor{red}{\textbf{(}$\textbf{2.18}$\textbf{)}} & $\textcolor{red}{\textbf{74.81}}$ \textcolor{red}{\textbf{(}$\textbf{1.02}$\textbf{)}} & $\textbf{85.93}$ \textbf{(}$\textbf{1.02}$\textbf{)} & $\textcolor{red}{\textbf{68.98}}$ \textcolor{red}{\textbf{(}$\textbf{1.14}$\textbf{)}} & $\textcolor{red}{\textbf{80.64}}$ \textcolor{red}{\textbf{(}$\textbf{2.94}$\textbf{)}} \\
        \midrule
        \multirow{2}{*}{LP-FT} & $1$ ($3$) & $\textcolor{red}{\textbf{90.99}}$ \textcolor{red}{\textbf{(}$\textbf{1.79}$\textbf{)}} & $\textbf{67.11}$ \textbf{(}$\textbf{2.44}$\textbf{)} & $82.93$ ($1.40$) & $\textcolor{red}{\textbf{63.37}}$ \textcolor{red}{\textbf{(}$\textbf{4.66}$\textbf{)}} & $\textcolor{red}{\textbf{75.56}}$ \textcolor{red}{\textbf{(}$\textbf{3.39}$\textbf{)}} \\
        & $3$ ($10$) & $\textcolor{red}{\textbf{93.67}}$ \textcolor{red}{\textbf{(}$\textbf{1.23}$\textbf{)}} & $\textcolor{red}{\textbf{75.63}}$ \textcolor{red}{\textbf{(}$\textbf{3.29}$\textbf{)}} & $\textbf{86.70}$ \textbf{(}$\textbf{0.30}$\textbf{)} & $\textcolor{red}{\textbf{72.02}}$ \textcolor{red}{\textbf{(}$\textbf{3.82}$\textbf{)}} & $\textcolor{red}{\textbf{81.83}}$ \textcolor{red}{\textbf{(}$\textbf{1.33}$\textbf{)}} \\
        \bottomrule
    \end{tabular}
    }
    \end{center}
    \caption{Average performance and standard deviation of SFUDA and few-shot fine-tuning methods under OoD and imbalance scenarios over three random runs. Notations are the same as in Table~\ref{tab:main}.}
    \label{tab:pitfalls}
\end{table}

\subsection{Why Few-shot Fine-tuning Works}\label{sec:imgnet_exp}
The overfitting problem can emerge when attempting to train high-capacity models with a limited amount of training data, resulting in poor generalization performance on unseen data. It is known that DNNs possess a high learning capacity, allowing them to perfectly fit even random Gaussian noise images during training~\cite{zhang2017understanding}. Hence, it is reasonable to expect that severe overfitting may occur when only a few training data are available. However, we observe that few-shot find-tuning does not suffer from overfitting, as shown in Table~\ref{tab:main}.

We assume that this phenomenon arises from the shared semantic information between the source and target domain (i.e., identical label spaces $\mathcal{Y}_s = \mathcal{Y}_t$). To test our hypothesis, we consider fine-tuning the ImageNet-1k~\cite{deng2009imagenet} pretrained model to the target domain using OfficeHome, to investigate the effect of label space differences between the source and target domains on performance changes. If the relationship between the source and target domain has no impact on mitigating overfitting, then the ImageNet pretrained model will show similar tendencies to the source pretrained model for few-shot SFDA. We employ three different fine-tuning strategies: linear probing (LP, training only a linear classifier), FT, and LP-FT. For both pretrained models, we use the same experimental setting as described in Sec~\ref{sec:4.3}. 

As shown in Table~\ref{tab:imagenet}, the few-shot adaptation performance of the ImageNet pretrained model and source pretrained model are significantly different. We observe that the ImageNet pretrained model is clearly prone to overfitting from the results that LP performs better than FT in both $1$-shot and $3$-shot configurations. LP-FT shows some improvement over FT, but it does not surpass the performance of LP. On the other hand, when adapting the source pretrained model, FT consistently performs better than LP even though the number of learnable parameters is incomparably larger. LP-FT shows further improvement in $3$-shot configuration. Moreover, the performance is improved over the ``No adapt'' baseline in all configurations when using the source pretrained model. These results indicate that the relevance of semantics between source and target domains alleviates the overfitting problem caused by few-shot training, supporting our experimental results that few-shot fine-tuning works surprisingly well for SFDA.

\begin{table}[t!]
    \begin{center}
    \resizebox{0.48\textwidth}{!}{%
    \begin{tabular}{l|c|ccc}
        \toprule
        \textbf{Pretrained} & \textbf{$\#$ shot} & \textbf{LP} & \textbf{FT} & \textbf{LP-FT} \\
        \midrule
        \multirow{2}{*}{ImageNet} & $1$ & $42.57$ $(0.56)$ & $31.71$ $(1.14)$ & $42.56$ $(0.65)$ \\
        & $3$ & $60.67$ $(0.91)$ & $53.43$ $(0.54)$ & $60.34$ $(0.82)$ \\
        \midrule
        \multirow{3}{*}{Source} & $1$ & $62.34$ $(0.62)$ & $64.45$ $(0.94)$ & $63.63$ $(1.78)$ \\
        & $3$ & $66.11$ $(0.29)$ & $70.38$ $(0.26)$ & $71.94$ $(1.10)$ \\
        \cmidrule{2-5}
        & No adapt & \multicolumn{3}{c}{$59.26$ $(0.57)$} \\
        \bottomrule
    \end{tabular} 
    }
    \end{center}
    \caption{Average accuracy and standard deviation of LP, FT, and LP-FT when fine-tuning ImageNet pretrained and source domain pretrained models for SFDA, respectively.}
    \label{tab:imagenet}
\end{table}

\subsection{Comparison with Transfer Learning Methods}\label{sec:compare_tl}
From the perspective of transferring pretrained features to the target domain with a few labeled data, advanced transfer learning methods can be considered as a potential solution. There are several transfer learning methods that have shown to be effective in small data regimes, such as L2-SP~\cite{xuhong2018explicit}, DELTA~\cite{li2019delta}, and BSS~\cite{chen2019catastrophic}. Additionally, the few-shot SFDA method, LCCS~\cite{zhang2022few}, can also be considered as a parameter-efficient fine-tuning method designed for SFDA that only trains batch normalization and the final classification layers.
To demonstrate that the few-shot SFDA methods, FT and LP-FT, are viable solutions, we compare FT and LP-FT with the above methods on the Office31 and OfficeHome datasets under $1$, $3$, and $5$-shot settings. For a fair comparison, we employ the same source pretrained model used in the previous experiments for all methods. The hyperparameters are chosen through $1$-shot validation for transfer learning methods, and for LCCS, we use the same hyperparameters presented in the corresponding paper.

Table~\ref{tab:transfer} summarizes the comparison results. It is observed that none of the comparison methods outperforms the simplest FT and LP-FT. The best performance is achieved by LP-FT in four out of six cases. 
These results also support our claim that few-shot fine-tuning methods, such as FT and LP-FT, can achieve satisfactory performance without the need for complicated techniques in SFDA problems.

\begin{table}[t!]
    \begin{center}
    \resizebox{0.48\textwidth}{!}{%
    \begin{tabular}{p{60pt}|ccc|ccc}
        \toprule
        & \multicolumn{3}{c|}{\textbf{Office31}} & \multicolumn{3}{c}{\textbf{OfficeHome}} \\
        & $1$-shot & $3$-shot & $5$-shot & $1$-shot & $3$-shot & $5$-shot \\
        \midrule
        L2-SP~\cite{xuhong2018explicit} & $82.99$ & $87.43$ & $90.20$ & $64.00$ & $70.13$ & $73.34$ \\
        DELTA~\cite{li2019delta} & $\textbf{83.41}$ & $87.66$ & $90.55$ & $\textbf{64.50}$ & $70.19$ & $72.93$ \\
        BSS~\cite{chen2019catastrophic} & $82.87$ & $87.26$ & $90.26$ & $63.91$ & $70.20$ & $73.29$ \\
        LCCS~\cite{zhang2022few} & $79.16$ & $85.75$ & $89.17$ & $60.53$ & $65.54$ & $68.18$ \\
        \midrule
        FT & $82.88$ & $87.55$ & $90.07$ & $64.45$ & $70.38$ & $73.37$ \\
        LP-FT & $82.85$ & $\textbf{88.93}$ & $\textbf{90.93}$ & $63.63$ & $\textbf{71.94}$ & $\textbf{74.55}$ \\
        \bottomrule
    \end{tabular}
    }
    \end{center}
    \caption{Average accuracy over three random runs for comparison of transfer learning methods, LCCS, FT, and LP-FT. \textbf{Bold} denotes the best performance on each few-shot configuration.}
    \label{tab:transfer}
\end{table}

\section{Conclusion}
In this work, we address the practicality of SFUDA and shed a light on the pitfalls often overlooked in SFUDA caused by depending solely on unlabeled target data. We demonstrate that these pitfalls, including a lack of principled ways to select optimal hyperparameters and scenarios where the ideal assumptions for the unlabeled target dataset do not hold, significantly limit the practical applicability of existing SFUDA methods. We argue that fine-tuning the source pretrained model with a few-shot labeled data is a more viable solution that inherently overcomes these limitations. Notably, our findings demonstrate that carefully fine-tuned source pretrained models do not suffer from overfitting with negligible sampling bias, despite being trained under the few-shot setting. Moreover, we show that performance can be further enhanced by learning a classifier head before fine-tuning, known as LP-FT. Further investigation into other regularization techniques that may be useful for few-shot SFDA would be of interest.

\section*{Acknowledgement}
This work was supported by the National Research Foundation of Korea (NRF) grant funded by the Korea government (MSIT) (NRF-2021R1C1C1011907).

{\small
\bibliographystyle{ieee_fullname}
\bibliography{egbib}

\begin{thebibliography}{10}\itemsep=-1pt

\bibitem{beery2018recognition}
Sara Beery, Grant Van~Horn, and Pietro Perona.
\newblock Recognition in terra incognita.
\newblock In {\em Proceedings of the European Conference on Computer Vision},
  2018.

\bibitem{berthelot2019mixmatch}
David Berthelot, Nicholas Carlini, Ian Goodfellow, Nicolas Papernot, Avital
  Oliver, and Colin~A Raffel.
\newblock Mixmatch: A holistic approach to semi-supervised learning.
\newblock In {\em Advances in Neural Information Processing Systems},
  volume~32, 2019.

\bibitem{branco2016survey}
Paula Branco, Lu\'{\i}s Torgo, and Rita~P. Ribeiro.
\newblock A survey of predictive modeling on imbalanced domains.
\newblock {\em ACM Computing Surveys (CSUR)}, 49(2):1--50, 2016.

\bibitem{chen2020deep}
Guangyi Chen, Yuhao Lu, Jiwen Lu, and Jie Zhou.
\newblock Deep credible metric learning for unsupervised domain adaptation
  person re-identification.
\newblock In Andrea Vedaldi, Horst Bischof, Thomas Brox, and Jan-Michael Frahm,
  editors, {\em Computer Vision -- ECCV 2020}, pages 643--659, 2020.

\bibitem{chen2019catastrophic}
Xinyang Chen, Sinan Wang, Bo Fu, Mingsheng Long, and Jianmin Wang.
\newblock Catastrophic forgetting meets negative transfer: Batch spectral
  shrinkage for safe transfer learning.
\newblock {\em Advances in Neural Information Processing Systems}, 32, 2019.

\bibitem{choi2010exploiting}
Myung~Jin Choi, Joseph~J. Lim, Antonio Torralba, and Alan~S. Willsky.
\newblock Exploiting hierarchical context on a large database of object
  categories.
\newblock In {\em 2010 IEEE Computer Society Conference on Computer Vision and
  Pattern Recognition}, pages 129--136, 2010.

\bibitem{deng2009imagenet}
Jia Deng, Wei Dong, Richard Socher, Li-Jia Li, Kai Li, and Li Fei-Fei.
\newblock Imagenet: A large-scale hierarchical image database.
\newblock In {\em 2009 IEEE Conference on Computer Vision and Pattern
  Recognition}, pages 248--255, 2009.

\bibitem{everingham2009pascal}
Mark Everingham, Luc~Van Gool, Christopher K.~I. Williams, John Winn, and
  Andrew Zisserman.
\newblock The pascal visual object classes (voc) challenge.
\newblock {\em International Journal of Computer Vision}, 88:303--308, 2009.

\bibitem{fang2013unbiased}
Chen Fang, Ye Xu, and Daniel~N. Rockmore.
\newblock Unbiased metric learning: On the utilization of multiple datasets and
  web images for softening bias.
\newblock In {\em Proceedings of the IEEE International Conference on Computer
  Vision}, pages 1657--1664, 2013.

\bibitem{fang2022source}
Yuqi Fang, Pew-Thian Yap, Weili Lin, Hongtu Zhu, and Mingxia Liu.
\newblock Source-free unsupervised domain adaptation: A survey.
\newblock {\em arXiv preprint arXiv:2301.00265}, 2022.

\bibitem{li2004learning}
Li Fei-Fei, R. Fergus, and P. Perona.
\newblock Learning generative visual models from few training examples: An
  incremental bayesian approach tested on 101 object categories.
\newblock In {\em 2004 Conference on Computer Vision and Pattern Recognition
  Workshop}, pages 178--178, 2004.

\bibitem{foret2021sharpnessaware}
Pierre Foret, Ariel Kleiner, Hossein Mobahi, and Behnam Neyshabur.
\newblock Sharpness-aware minimization for efficiently improving
  generalization.
\newblock In {\em International Conference on Learning Representations}, 2021.

\bibitem{ganin2016domain}
Yaroslav Ganin, Evgeniya Ustinova, Hana Ajakan, Pascal Germain, Hugo
  Larochelle, Fran{\c{c}}ois Laviolette, Mario Marchand, and Victor Lempitsky.
\newblock Domain-adversarial training of neural networks.
\newblock {\em The Journal of Machine Learning Research}, 17(1):2096--2030,
  2016.

\bibitem{gulrajani2021in}
Ishaan Gulrajani and David Lopez-Paz.
\newblock In search of lost domain generalization.
\newblock In {\em International Conference on Learning Representations}, 2021.

\bibitem{he2016deep}
Kaiming He, Xiangyu Zhang, Shaoqing Ren, and Jian Sun.
\newblock Deep residual learning for image recognition.
\newblock In {\em Proceedings of the IEEE Conference on Computer Vision and
  Pattern Recognition}, pages 770--778, 2016.

\bibitem{ioffe2015batch}
Sergey Ioffe and Christian Szegedy.
\newblock Batch normalization: Accelerating deep network training by reducing
  internal covariate shift.
\newblock In {\em International conference on machine learning}, pages
  448--456, 2015.

\bibitem{iwasawa2021test}
Yusuke Iwasawa and Yutaka Matsuo.
\newblock Test-time classifier adjustment module for model-agnostic domain
  generalization.
\newblock In {\em Advances in Neural Information Processing Systems},
  volume~34, pages 2427--2440, 2021.

\bibitem{johnson2019survey}
Justin~M. Johnson and Taghi~M. Khoshgoftaar.
\newblock Survey on deep learning with class imbalance.
\newblock {\em Journal of Big Data}, 6(1):1--54, 2019.

\bibitem{kingma2015adam}
Diederik~P. Kingma and Jimmy Ba.
\newblock Adam: A method for stochastic optimization.
\newblock In {\em International Conference on Learning Representations}, 2015.

\bibitem{kumar2022fine}
Ananya Kumar, Aditi Raghunathan, Robbie~Matthew Jones, Tengyu Ma, and Percy
  Liang.
\newblock Fine-tuning can distort pretrained features and underperform
  out-of-distribution.
\newblock In {\em International Conference on Learning Representations}, 2022.

\bibitem{kundu2020universal}
Jogendra~Nath Kundu, Naveen Venkat, Rahul~M V, and R.~Venkatesh Babu.
\newblock Universal source-free domain adaptation.
\newblock In {\em Proceedings of the IEEE/CVF Conference on Computer Vision and
  Pattern Recognition}, pages 4544--4553, 2020.

\bibitem{kurmi2021domain}
Vinod~K. Kurmi, Venkatesh~K. Subramanian, and Vinay~P. Namboodiri.
\newblock Domain impression: A source data free domain adaptation method.
\newblock In {\em Proceedings of the IEEE/CVF Winter Conference on Applications
  of Computer Vision}, pages 615--625, 2021.

\bibitem{lee2022confidence}
Jonghyun Lee, Dahuin Jung, Junho Yim, and Sungroh Yoon.
\newblock Confidence score for source-free unsupervised domain adaptation.
\newblock In {\em Proceedings of the 39th International Conference on Machine
  Learning}, volume 162, pages 12365--12377, 2022.

\bibitem{li2020model}
Rui Li, Qianfen Jiao, Wenming Cao, Hau-San Wong, and Si Wu.
\newblock Model adaptation: Unsupervised domain adaptation without source data.
\newblock In {\em Proceedings of the IEEE/CVF Conference on Computer Vision and
  Pattern Recognition}, pages 9641--9650, June 2020.

\bibitem{li2021transferable}
Shuang Li, Mixue Xie, Kaixiong Gong, Chi~Harold Liu, Yulin Wang, and Wei Li.
\newblock Transferable semantic augmentation for domain adaptation.
\newblock In {\em Proceedings of the IEEE/CVF Conference on Computer Vision and
  Pattern Recognition}, pages 11516--11525, 2021.

\bibitem{li2021imbalanced}
Xinhao Li, Jingjing Li, Lei Zhu, Guoqing Wang, and Zi Huang.
\newblock Imbalanced source-free domain adaptation.
\newblock In {\em Proceedings of the 29th ACM International Conference on
  Multimedia}, pages 3330--3339, 2021.

\bibitem{li2019delta}
Xingjian Li, Haoyi Xiong, Hanchao Wang, Yuxuan Rao, Liping Liu, and Jun Huan.
\newblock {DELTA}: Deep learning transfer using feature map with attention for
  convolutional networks.
\newblock In {\em International Conference on Learning Representations}, 2019.

\bibitem{liang2020we}
Jian Liang, Dapeng Hu, and Jiashi Feng.
\newblock Do we really need to access the source data? {S}ource hypothesis
  transfer for unsupervised domain adaptation.
\newblock In {\em Proceedings of the 37th International Conference on Machine
  Learning}, volume 119, pages 6028--6039, 2020.

\bibitem{liang2021umad}
Jian Liang, Dapeng Hu, Jiashi Feng, and Ran He.
\newblock Umad: Universal model adaptation under domain and category shift.
\newblock {\em arXiv preprint arXiv:2112.08553}, 2021.

\bibitem{liang2021source}
Jian Liang, Dapeng Hu, Yunbo Wang, Ran He, and Jiashi Feng.
\newblock Source data-absent unsupervised domain adaptation through hypothesis
  transfer and labeling transfer.
\newblock {\em IEEE Transactions on Pattern Analysis and Machine Intelligence},
  44(11):8602--8617, 2022.

\bibitem{mekhazni2020unsupervised}
Djebril Mekhazni, Amran Bhuiyan, George Ekladious, and Eric Granger.
\newblock Unsupervised domain adaptation in the dissimilarity space for person
  re-identification.
\newblock In Andrea Vedaldi, Horst Bischof, Thomas Brox, and Jan-Michael Frahm,
  editors, {\em Computer Vision -- ECCV 2020}, pages 159--174, 2020.

\bibitem{nado2020evaluating}
Zachary Nado, Shreyas Padhy, D Sculley, Alexander D'Amour, Balaji
  Lakshminarayanan, and Jasper Snoek.
\newblock Evaluating prediction-time batch normalization for robustness under
  covariate shift.
\newblock {\em arXiv preprint arXiv:2006.10963}, 2020.

\bibitem{niu2023towards}
Shuaicheng Niu, Jiaxiang Wu, Yifan Zhang, Zhiquan Wen, Yaofo Chen, Peilin Zhao,
  and Mingkui Tan.
\newblock Towards stable test-time adaptation in dynamic wild world.
\newblock In {\em The Eleventh International Conference on Learning
  Representations}, 2023.

\bibitem{oza2021unsupervised}
Poojan Oza, Vishwanath~A Sindagi, Vibashan VS, and Vishal~M Patel.
\newblock Unsupervised domain adaptation of object detectors: A survey.
\newblock {\em arXiv preprint arXiv:2105.13502}, 2021.

\bibitem{pandey2020unsupervised}
Prashant Pandey, Aayush~Kumar Tyagi, Sameer Ambekar, and A.~P. Prathosh.
\newblock Unsupervised domain adaptation for semantic segmentation of nir
  images through generative latent search.
\newblock In Andrea Vedaldi, Horst Bischof, Thomas Brox, and Jan-Michael Frahm,
  editors, {\em Computer Vision -- ECCV 2020}, pages 413--429, 2020.

\bibitem{peng2017visda}
Xingchao Peng, Ben Usman, Neela Kaushik, Judy Hoffman, Dequan Wang, and Kate
  Saenko.
\newblock Visda: The visual domain adaptation challenge.
\newblock {\em arXiv preprint arXiv:1710.06924}, 2017.

\bibitem{ijcai2021p0402}
Zhen Qiu, Yifan Zhang, Hongbin Lin, Shuaicheng Niu, Yanxia Liu, Qing Du, and
  Mingkui Tan.
\newblock Source-free domain adaptation via avatar prototype generation and
  adaptation.
\newblock In {\em Proceedings of the Thirtieth International Joint Conference
  on Artificial Intelligence}, pages 2921--2927, 2021.

\bibitem{quinonero2008dataset}
Joaquin Quinonero-Candela, Masashi Sugiyama, Anton Schwaighofer, and Neil~D
  Lawrence.
\newblock {\em Dataset shift in machine learning}.
\newblock 2008.

\bibitem{russell2008label}
Bryan~C. Russell, Antonio Torralba, Kevin~P. Murphy, and William~T. Freeman.
\newblock Labelme: A database and web-based tool for image annotation.
\newblock {\em International Journal of Computer Vision}, 77:157--173, 2008.

\bibitem{saenko2010adapting}
Kate Saenko, Brian Kulis, Mario Fritz, and Trevor Darrell.
\newblock Adapting visual category models to new domains.
\newblock In Kostas Daniilidis, Petros Maragos, and Nikos Paragios, editors,
  {\em Computer Vision -- ECCV 2010}, pages 213--226, 2010.

\bibitem{saito2020universal}
Kuniaki Saito, Donghyun Kim, Stan Sclaroff, and Kate Saenko.
\newblock Universal domain adaptation through self supervision.
\newblock In {\em Advances in Neural Information Processing Systems},
  volume~33, pages 16282--16292, 2020.

\bibitem{saito2021tune}
Kuniaki Saito, Donghyun Kim, Piotr Teterwak, Stan Sclaroff, Trevor Darrell, and
  Kate Saenko.
\newblock Tune it the right way: Unsupervised validation of domain adaptation
  via soft neighborhood density.
\newblock In {\em Proceedings of the IEEE/CVF International Conference on
  Computer Vision}, pages 9184--9193, 2021.

\bibitem{salimans2016weight}
Tim Salimans and Durk~P Kingma.
\newblock Weight normalization: A simple reparameterization to accelerate
  training of deep neural networks.
\newblock {\em Advances in neural information processing systems}, 29, 2016.

\bibitem{shi2022pairwise}
Weili Shi, Ronghang Zhu, and Sheng Li.
\newblock Pairwise adversarial training for unsupervised class-imbalanced
  domain adaptation.
\newblock In {\em Proceedings of the 28th ACM SIGKDD Conference on Knowledge
  Discovery and Data Mining}, pages 1598--1606, 2022.

\bibitem{sun2022source}
Wujie Sun, Qi Chen, Can Wang, Deshi Ye, and Chun Chen.
\newblock Source-free unsupervised domain adaptation in imbalanced datasets.
\newblock In {\em 2022 5th International Conference on Data Science and
  Information Technology}, pages 1--6, 2022.

\bibitem{tan2020class}
Shuhan Tan, Xingchao Peng, and Kate Saenko.
\newblock Class-imbalanced domain adaptation: An empirical odyssey.
\newblock In {\em Computer Vision -- ECCV 2020 Workshops}, pages 585--602,
  2020.

\bibitem{tzeng2017adversarial}
Eric Tzeng, Judy Hoffman, Kate Saenko, and Trevor Darrell.
\newblock Adversarial discriminative domain adaptation.
\newblock In {\em Proceedings of the IEEE Conference on Computer Vision and
  Pattern Recognition}, pages 7167--7176, 2017.

\bibitem{van2008visualizing}
Laurens Van~der Maaten and Geoffrey Hinton.
\newblock Visualizing data using t-sne.
\newblock {\em Journal of machine learning research}, 9(11), 2008.

\bibitem{venkateswara2017deep}
Hemanth Venkateswara, Jose Eusebio, Shayok Chakraborty, and Sethuraman
  Panchanathan.
\newblock Deep hashing network for unsupervised domain adaptation.
\newblock In {\em Proceedings of the IEEE Conference on Computer Vision and
  Pattern Recognition}, pages 5018--5027, 2017.

\bibitem{vs2021mega}
Vibashan VS, Vikram Gupta, Poojan Oza, Vishwanath~A. Sindagi, and Vishal~M.
  Patel.
\newblock Mega-cda: Memory guided attention for category-aware unsupervised
  domain adaptive object detection.
\newblock In {\em Proceedings of the IEEE/CVF Conference on Computer Vision and
  Pattern Recognition}, pages 4516--4526, 2021.

\bibitem{wang2020tent}
Dequan Wang, Evan Shelhamer, Shaoteng Liu, Bruno Olshausen, and Trevor Darrell.
\newblock Tent: Fully test-time adaptation by entropy minimization.
\newblock In {\em International Conference on Learning Representations}, 2021.

\bibitem{wang2020classes}
Haoran Wang, Tong Shen, Wei Zhang, Ling-Yu Duan, and Tao Mei.
\newblock Classes matter: A fine-grained adversarial approach to cross-domain
  semantic segmentation.
\newblock In Andrea Vedaldi, Horst Bischof, Thomas Brox, and Jan-Michael Frahm,
  editors, {\em Computer Vision -- ECCV 2020}, pages 642--659, 2020.

\bibitem{wang2016training}
Shoujin Wang, Wei Liu, Jia Wu, Longbing Cao, Qinxue Meng, and Paul~J Kennedy.
\newblock Training deep neural networks on imbalanced data sets.
\newblock In {\em 2016 International Joint Conference on Neural Networks},
  pages 4368--4374. IEEE, 2016.

\bibitem{wang2020self}
Shanshan Wang and Lei Zhang.
\newblock Self-adaptive re-weighted adversarial domain adaptation.
\newblock In {\em Proceedings of the 29th International Joint Conference on
  Artificial Intelligence}, number 440, pages 3181--3187, 2021.

\bibitem{xuhong2018explicit}
LI Xuhong, Yves Grandvalet, and Franck Davoine.
\newblock Explicit inductive bias for transfer learning with convolutional
  networks.
\newblock In {\em International Conference on Machine Learning}, pages
  2825--2834, 2018.

\bibitem{yang2021exploring}
Jinyu Yang, Chunyuan Li, Weizhi An, Hehuan Ma, Yuzhi Guo, Yu Rong, Peilin Zhao,
  and Junzhou Huang.
\newblock Exploring robustness of unsupervised domain adaptation in semantic
  segmentation.
\newblock In {\em Proceedings of the IEEE/CVF International Conference on
  Computer Vision}, pages 9194--9203, 2021.

\bibitem{yang2021exploiting}
Shiqi Yang, yaxing wang, Joost van~de Weijer, Luis Herranz, and Shangling Jui.
\newblock Exploiting the intrinsic neighborhood structure for source-free
  domain adaptation.
\newblock In {\em Advances in Neural Information Processing Systems},
  volume~34, pages 29393--29405, 2021.

\bibitem{yang2021generalized}
Shiqi Yang, Yaxing Wang, Joost van~de Weijer, Luis Herranz, and Shangling Jui.
\newblock Generalized source-free domain adaptation.
\newblock In {\em Proceedings of the IEEE/CVF International Conference on
  Computer Vision}, pages 8978--8987, 2021.

\bibitem{yang2022attracting}
Shiqi Yang, Yaxing Wang, Kai Wang, SHANGLING JUI, and Joost van~de weijer.
\newblock Attracting and dispersing: A simple approach for source-free domain
  adaptation.
\newblock In {\em Advances in Neural Information Processing Systems}, 2022.

\bibitem{yang2022one}
Shiqi Yang, Yaxing Wang, Kai Wang, Shangling Jui, and Joost van~de Weijer.
\newblock One ring to bring them all: Towards open-set recognition under domain
  shift.
\newblock {\em arXiv preprint arXiv:2206.03600}, 2022.

\bibitem{zhang2017understanding}
Chiyuan Zhang, Samy Bengio, Moritz Hardt, Benjamin Recht, and Oriol Vinyals.
\newblock Understanding deep learning requires rethinking generalization.
\newblock In {\em International Conference on Learning Representations}, 2017.

\bibitem{zhang2022few}
Wenyu Zhang, Li Shen, Wanyue Zhang, and Chuan-Sheng Foo.
\newblock Few-shot adaptation of pre-trained networks for domain shift.
\newblock In {\em Proceedings of the 31st International Joint Conference on
  Artificial Intelligence}, pages 1665--1671, 2022.

\end{thebibliography}
}

\newpage
\onecolumn
\setcounter{figure}{0}
\setcounter{table}{0}
\renewcommand\thefigure{S\arabic{figure}}
\renewcommand\thetable{S\arabic{table}}
\renewcommand*{\theHtable}{\thetable}
\renewcommand*{\theHfigure}{\thefigure}

\appendix

\section{Reproduction Results of SFUDA Methods}
A common practice in current SFUDA methods is to use the entire target domain data for both training and testing. In contrast, we split the target domain data into training and test sets to facilitate a more rigorous comparison. To verify the validity of our experimental results, we present our reproduced results on the OfficeHome dataset following the standard SFUDA practice (i.e., using the entire target data for both training and testing), along with the performance reported in the corresponding paper. We also provide the results with train/test splits. All the results are produced using the official codes released by the corresponding authors. The hyperparameters for each method are set to the same values as those reported in the literature or official codes.

As demonstrated in Table~\ref{tab:reproduce}, the performances of SFUDA methods are successfully reproduced with no considerable margin compared to the reported values. As expected, the SFUDA methods show slightly degraded performance under the rigorous setting, i.e., train/test splitting. This indicates that current SFUDA methods may not achieve the expected level of generalization performance when applied to the unseen target domain data.

\begin{table}[H]
    \begin{center}
    \resizebox{1\textwidth}{!}{%
    \begin{tabular}{l|cccccccccccccc}
        \toprule
        Method & \multicolumn{1}{c}{Ar→Cl} & \multicolumn{1}{c}{Ar→Pr} & \multicolumn{1}{c}{Ar→Rw} & \multicolumn{1}{c}{Cl→Ar} & \multicolumn{1}{c}{Cl→Pr} & \multicolumn{1}{c}{Cl→Rw} & \multicolumn{1}{c}{Pr→Ar} & \multicolumn{1}{c}{Pr→Cl} & \multicolumn{1}{c}{Pr→Rw} & \multicolumn{1}{c}{Rw→Ar} & \multicolumn{1}{c}{Rw→Cl} & \multicolumn{1}{c}{Rw→Pr} & \multicolumn{1}{c}{Avg} \\
        \midrule
        \multicolumn{1}{l|}{SHOT~\cite{liang2020we}} & $57.1$ & $78.1$ & $81.5$ & $68.0$ & $78.2$ & $78.1$ & $67.4$ & $54.9$ & $82.2$ & $73.3$ & $58.8$ & $84.3$ & $71.8$ \\
        \multicolumn{1}{l|}{SHOT~\cite{liang2020we}$^\dagger$} & $56.6$ & $78.2$ & $81.4$ & $68.0$ & $78.9$ & $78.0$ & $67.5$ & $54.6$ & $82.1$ & $73.5$ & $58.7$ & $83.7$ & $71.8$ \\
        \multicolumn{1}{l|}{SHOT~\cite{liang2020we}$^{\dagger\dagger}$} & $54.9$ & $75.5$ & $80.0$ & $64.7$ & $75.6$ & $77.1$ & $64.6$ & $53.7$ & $81.4$ & $71.7$ & $57.4$ & $82.3$ & $69.9$ \\
        \multicolumn{1}{l|}{SHOT$++$~\cite{liang2021source}} & $57.9$ & $79.7$ & $82.5$ & $68.5$ & $79.6$ & $79.3$ & $68.5$ & $57.0$ & $83.0$ & $73.7$ & $60.7$ & $84.9$ & $73.0$ \\
        \multicolumn{1}{l|}{SHOT$++$~\cite{liang2021source}$^\dagger$} & $58.1$ & $79.2$ & $82.4$ & $69.2$ & $80.2$ & $79.2$ & $68.8$ & $56.7$ & $82.6$ & $74.2$ & $60.0$ & $84.7$ & $72.9$ \\
        \multicolumn{1}{l|}{SHOT$++$~\cite{liang2021source}$^{\dagger\dagger}$} & $56.1$ & $77.8$ & $80.1$ & $65.6$ & $78.9$ & $77.4$ & $65.3$ & $53.6$ & $80.7$ & $71.3$ & $56.5$ & $83.6$ & $70.6$ \\
        \multicolumn{1}{l|}{AaD~\cite{yang2022attracting}} & $59.3$ & $79.3$ & $82.1$ & $68.9$ & $79.8$ & $79.5$ & $67.2$ & $57.4$ & $83.1$ & $72.1$ & $58.5$ & $85.4$ & $72.7$ \\
        \multicolumn{1}{l|}{AaD~\cite{yang2022attracting}$^\dagger$} & $58.7$ & $79.3$ & $81.3$ & $68.0$ & $79.9$ & $79.2$ & $66.9$ & $57.1$ & $82.4$ & $72.3$ & $58.5$ & $85.0$ & $72.4$ \\
        {AaD~\cite{yang2022attracting}$^{\dagger\dagger}$} & $56.2$ & $77.9$ & $79.7$ & $63.7$ & $80.0$ & $78.5$ & $63.1$ & $56.1$ & $80.8$ & $69.8$ & $57.9$ & $84.8$ & $70.7$ \\
        \multicolumn{1}{l|}{CoWA-JMDS~\cite{lee2022confidence}} & $56.9$ & $78.4$ & $81.0$ & $69.1$ & $80.0$ & $79.9$ & $67.7$ & $57.2$ & $82.4$ & $72.8$ & $60.5$ & $84.5$ & $72.5$ \\
        \multicolumn{1}{l|}{CoWA-JMDS~\cite{lee2022confidence}$^\dagger$} & $58.5$ & $78.6$ & $81.1$ & $68.2$ & $79.9$ & $79.5$ & $67.3$ & $57.0$ & $82.8$ & $72.3$ & $60.4$ & $84.4$ & $72.5$ \\
        \multicolumn{1}{l|}{CoWA-JMDS~\cite{lee2022confidence}$^{\dagger\dagger}$} & $57.5$ & $78.6$ & $81.5$ & $67.8$ & $79.4$ & $79.9$ & $65.8$ & $56.7$ & $82.3$ & $71.4$ & $59.9$ & $84.1$ & $72.1$ \\
        \multicolumn{1}{l|}{NRC~\cite{yang2021exploiting}} & $57.7$ & $80.3$ & $82.0$ & $68.1$ & $79.8$ & $78.6$ & $65.3$ & $56.4$ & $83.0$ & $71.0$ & $58.6$ & $85.6$ & $72.2$ \\
        \multicolumn{1}{l|}{NRC~\cite{yang2021exploiting}$^\dagger$} & $57.6$ & $79.2$ & $80.8$ & $67.6$ & $79.3$ & $79.0$ & $66.7$ & $56.7$ & $81.4$ & $72.7$ & $58.8$ & $85.2$ & $72.1$ \\
        {NRC~\cite{yang2021exploiting}$^{\dagger\dagger}$} & $55.6$ & $76.8$ & $79.7$ & $65.5$ & $76.8$ & $77.4$ & $65.1$ & $54.9$ & $80.5$ & $71.1$ & $57.0$ & $82.5$ & $70.3$ \\
        \multicolumn{1}{l|}{G-SFDA~\cite{yang2021generalized}} & $57.9$ & $78.6$ & $81.0$ & $66.7$ & $77.2$ & $77.2$ & $65.6$ & $56.0$ & $82.2$ & $72.0$ & $57.8$ & $83.4$ & $71.3$ \\
        \multicolumn{1}{l|}{G-SFDA~\cite{yang2021generalized}$^\dagger$} & $56.2$ & $76.4$ & $80.2$ & $65.8$ & $75.6$ & $77.2$ & $64.8$ & $53.9$ & $81.4$ & $72.0$ & $58.5$ & $84.0$ & $70.5$\\\multicolumn{1}{l|}{G-SFDA~\cite{yang2021generalized}$^{\dagger\dagger}$} & $55.0$ & $75.2$ & $79.8$ & $64.3$ & $73.7$ & $75.9$ & $63.4$ & $52.6$ & $80.6$ & $70.3$ & $57.6$ & $82.4$ & $69.2$ \\
        \bottomrule 
    \end{tabular} 
    }
    \end{center}
    \caption{Accuracies of SFUDA methods on OfficeHome. Ar, Cl, Pr, and Rw denote Art, Clipart, Product, and Real World, respectively. The results with no mark indicate the reported values in the literature. $^\dagger$ denotes the reproduced results under the standard SFUDA practice. $^{\dagger\dagger}$ denotes the results with train/test splits. All results are averaged over three random runs.}
    \label{tab:reproduce}
\end{table} 

\section{Label Distributions of Datasets}
\begin{figure}[H]
    \centering
    \subfigure[Office31]{
        \centering
        \includegraphics[width=0.25\columnwidth]{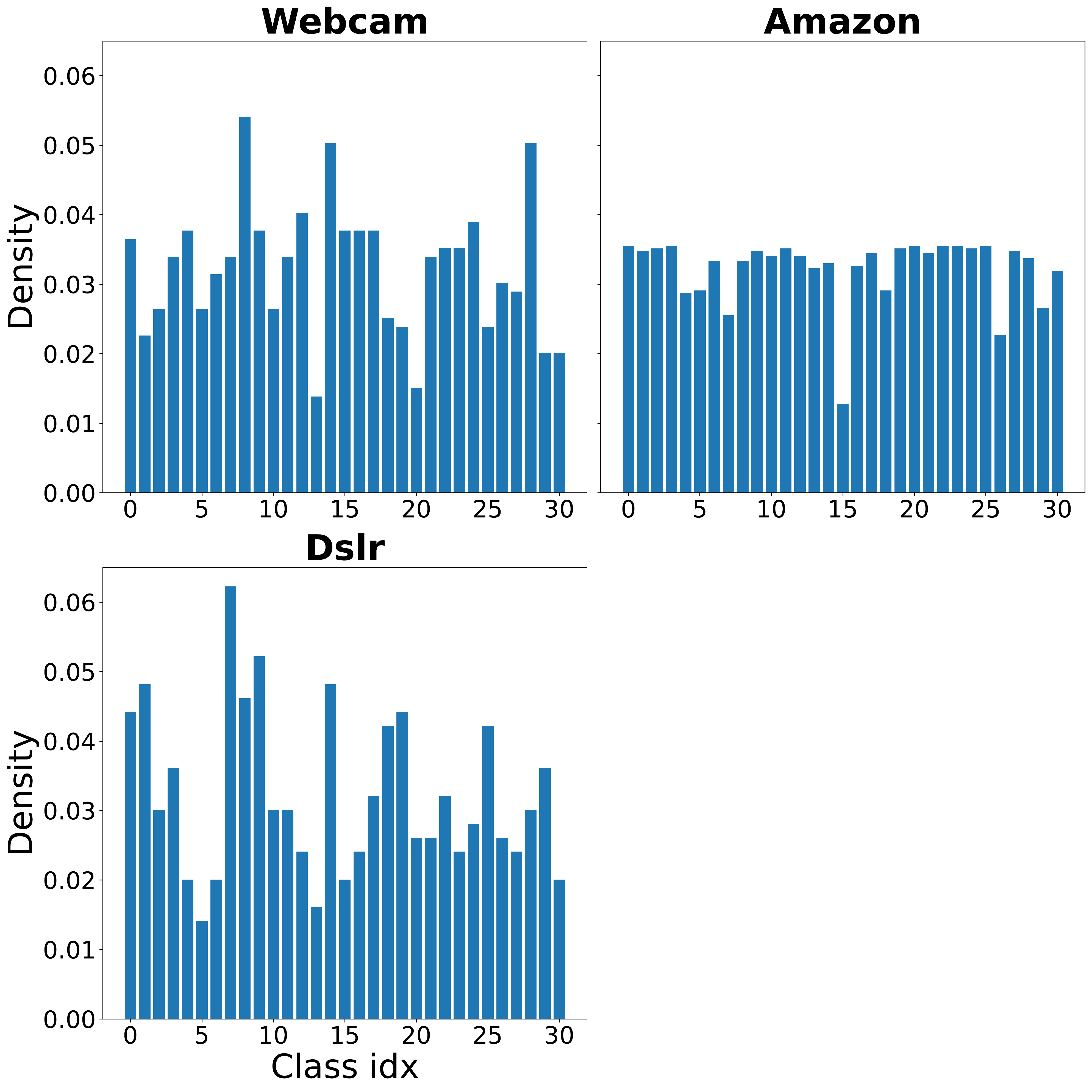}
        \label{fig:office31_dist}
        }
    \subfigure[VLCS]{
        \centering
        \includegraphics[width=0.25\columnwidth]{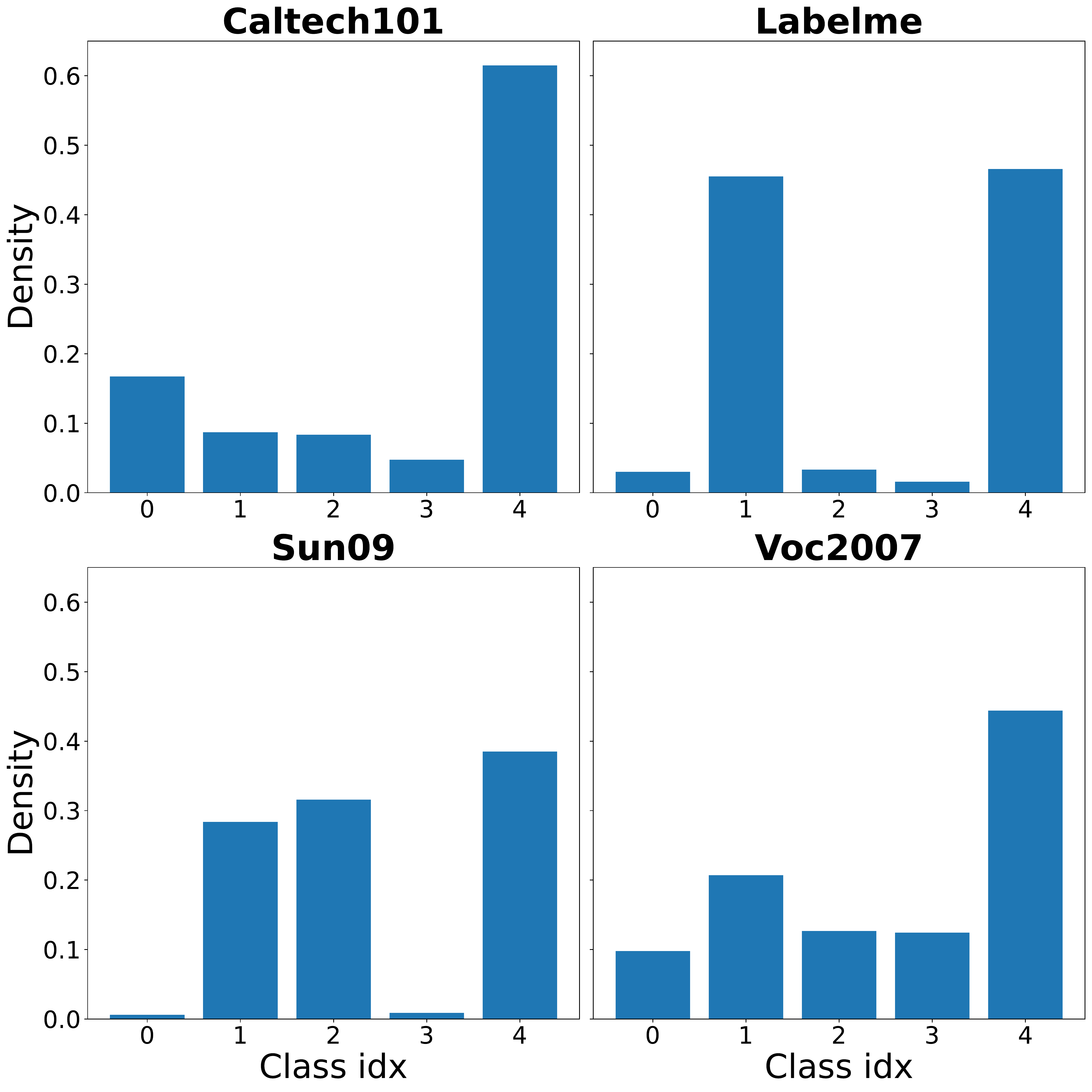}
        \label{fig:vlcs_dist}
        }
    \subfigure[Terra]{
        \centering
        \includegraphics[width=0.25\columnwidth]{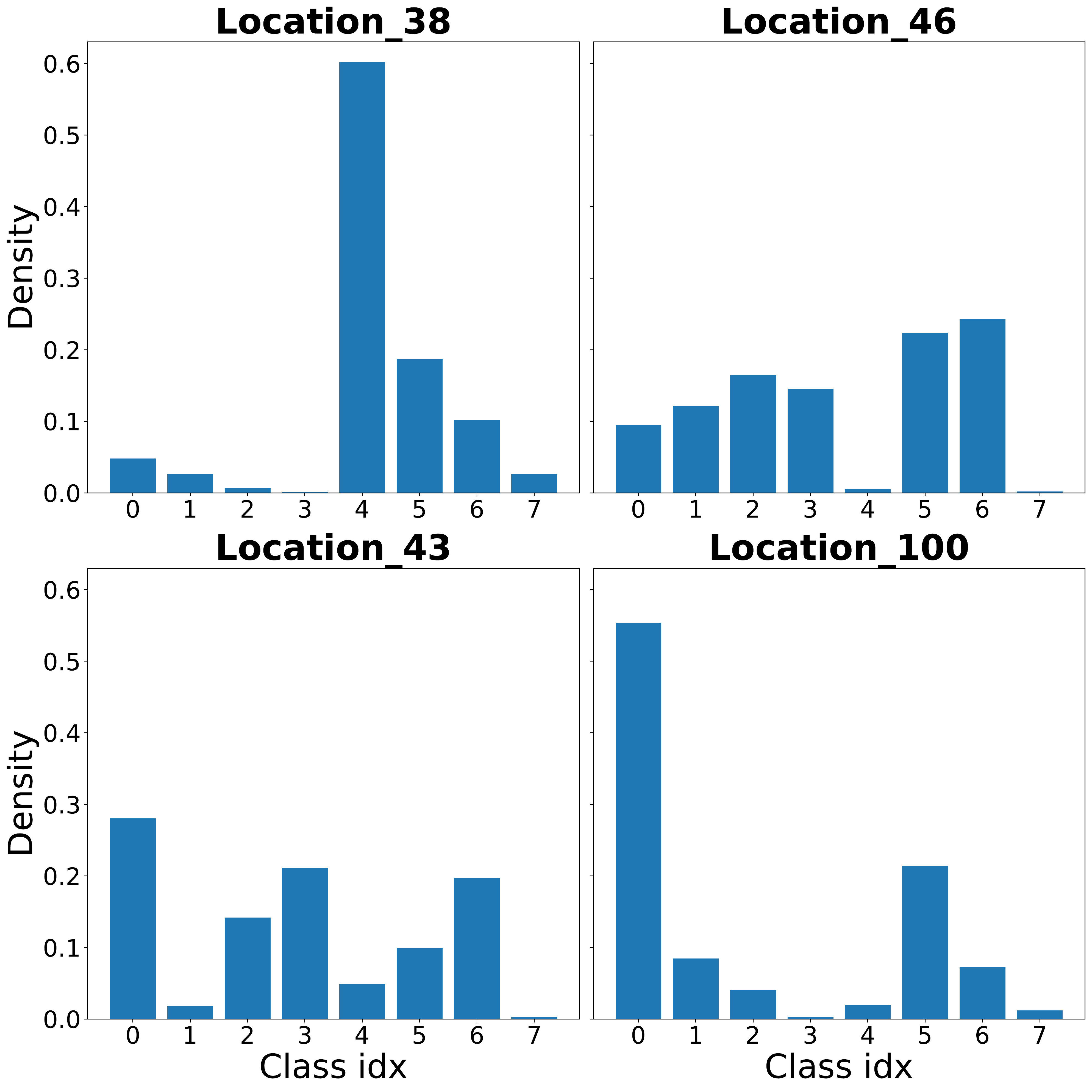}
        \label{fig:terra_dist}
        }
    \caption{Label distribution of each domain in Office31, VLCS, and Terra. In each plot, the $x$- and $y$-axes correspond to the class index and normalized frequency, respectively.}
    \label{fig:dist}
\end{figure}

Figure~\ref{fig:dist} presents the label distribution of each domain in three benchmark datasets used in this work. Office31, which is one of the most popular benchmarks in the SFUDA literature, has relatively balanced label distributions across all domains. However, there are substantial shifts in the label distributions in the VLCS and Terra datasets.

\section{Qualitative Results of Experiments in Section~\ref{sec:imgnet_exp}}
\begin{figure}[H]
    \centering
    \includegraphics[width=0.6\columnwidth]{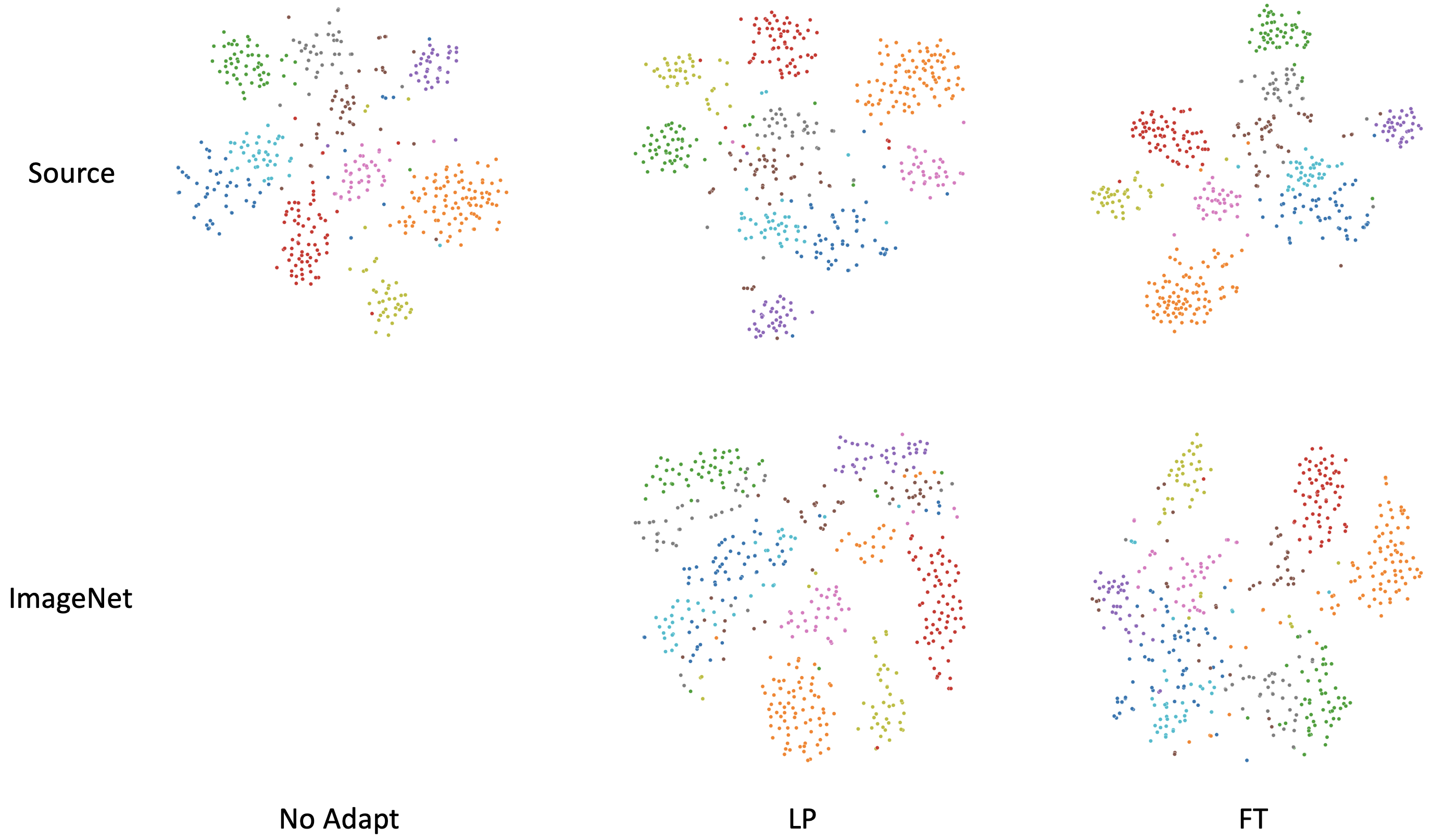}
    \caption{$t$-SNE~\cite{van2008visualizing} visualization results on the OfficeHome dataset (Rw $\rightarrow$ Pr). The first and second rows indicate results from the source pretrained model and ImageNet pretrained model, respectively. ``No Adapt'' for the ImageNet pretrained model is not considered since the dimension of the label space (i.e., $1,000$) is different from others. Different colors represent different classes. $10$ classes are randomly sampled for clear visualization.}
    \label{fig:tsne}
\end{figure}

In Section~\ref{sec:imgnet_exp}, we conduct a comparative analysis of the ImageNet pretrained model and source pretrained model to investigate the impact of label space discrepancies between the source and target domains. We provide qualitative results through $t$-SNE~\cite{van2008visualizing} visualization on the OfficeHome dataset. We consider the ImageNet pretrained model and source pretrained model, both adapted to the target domain with $1$-shot LP and $1$-shot FT. Also, we consider the source pretrained model without target adaptation. As shown in Figure~\ref{fig:tsne}, the source pretrained model successfully discriminates the target data, even without any target adaptation process. This implies that the semantic knowledge acquired from the source domain can be reasonably effective on the target domain. Also, FT of the source pretrained model differentiates the target data much more clearly than LP. In contrast, when using the ImageNet pretrained model, the features obtained through FT become more entangled than those of LP due to overfitting. These visualization results, in conjunction with the quantitative results presented in Section~\ref{sec:imgnet_exp}, support that the semantic relevance between the source and target domains can effectively mitigate overfitting even when using only a few samples for FT.

\section{Implementation Details for Sections~\ref{sec:imgnet_exp} and \ref{sec:compare_tl}}
To ensure clarity and reproducibility, we describe further implementation details for the ImageNet experiments in Section~\ref{sec:imgnet_exp} and transfer learning methods in Section~\ref{sec:compare_tl}.

\paragraph{ImageNet Experiments in Section~\ref{sec:imgnet_exp}}
For the ImageNet pretrained model, we use ResNet50 pretrained on ImageNet-1k as a backbone. Specifically, we replace the classification head with a new one that has the same number of output nodes as the number of classes in OfficeHome, which is 65. The hyperparameters used for few-shot training are identical to those employed for the source pre-trained model. We adapt the ImageNet pre-trained model to four OfficeHome domains, treating each one as the target domain, and report the average accuracy across the four target domains.

\paragraph{Transfer Learning Methods in Section~\ref{sec:compare_tl}}
For L2-SP, DELTA, and BSS, we search algorithm-specific hyperparameters through $1$-shot validation, which is also applied for few-shot FT. In L2-SP, there are two hyperparameters: the starting-point regularization coefficient $\alpha$ and the $L_2$ penalty of classification head $\beta$. The $\alpha$ is searched in $\{1e$-$02, 1e$-$03\}$ and the $\beta$ is searched in $\{0, 1e$-$03\}$. In DELTA, we only consider the coefficient of behavioral regularization $\delta$. The $\delta$ is searched in $\{1e$-$01,1e$-$02,1e$-$03,1e$-$04\}$. In BSS, there are two hyperparameters: the number of singular values $k$ and the trade-off parameter of BSS loss $\eta$. The $k$ is searched in $\{1,2\}$ and the $\eta$ is searched in $\{1e$-$03,1e$-$04\}$. The hyperparameters that are algorithm-agnostic, such as learning rate, batch size, etc., are set to the same values used in few-shot FT.

For LCCS, we use the same hyperparameters utilized for Office31 and OfficeHome experiments in the literature. However, the original source training procedure of LCCS differs from ours. We find that using our source model achieves slightly better performance compared to the performance reported for both Office31 and OfficeHome. According to the paper, the reported $5$-shot LCCS performances on Office31 and OfficeHome are $88.9$ and $67.8$, respectively. On the other hand, using our source model for $5$-shot LCCS achieves $89.17$ and $68.18$ on Office31 and OfficeHome, respectively.

\section{Detailed Experimental Results}
We provide the performance of SFUDA methods and few-shot SFDA methods across all source-target domain pairs. For comparison under vanilla SFDA settings (summarized in Table~\ref{tab:main}), Tables~\ref{tab:detail_o31},~\ref{tab:detail_oh},~\ref{tab:detail_visda},~\ref{tab:detail_vlcs}, and~\ref{tab:detail_terra} show the detailed results on Office31, OfficeHome, VisDA, VLCS, and Terra, respectively. For comparison under OoD scenarios (summarized in the left part of Table 4), Tables~\ref{tab:ood_o31},~\ref{tab:ood_oh}, and~\ref{tab:ood_visda} show the detailed results on Office31, OfficeHome, and VisDA, respectively. For comparison under label distribution shift (summarized in the right part of Table 4), we include the detailed results on OfficeHome (RSUT) and VisDA (RSUT) in Tables~\ref{tab:oh_rsut} and~\ref{tab:visda_rsut}, respectively. In Tables~\ref{tab:detail_visda},~\ref{tab:ood_visda}, and~\ref{tab:visda_rsut}, we report the accuracy for each class on the target domain since VisDA only has one source-target domain pair. All results are averaged over three random runs.

\begin{table}[H]
    \begin{center}
    \resizebox{0.5\textwidth}{!}{%
    \begin{tabular}{lc|cccccccc}
        \toprule
        Method & & \multicolumn{1}{c}{A→D} & \multicolumn{1}{c}{A→W} & \multicolumn{1}{c}{D→A} & \multicolumn{1}{c}{D→W} & \multicolumn{1}{c}{W→A} & \multicolumn{1}{c}{W→D} & \multicolumn{1}{c}{Avg}\\
        \midrule
        \multicolumn{2}{l|}{No adapt} & $77.33$ & $76.52$ & $57.33$ & $93.92$ & $62.00$ & $98.67$ & $77.63$ \\
        \multicolumn{2}{l|}{SHOT~\cite{liang2020we}} & $89.33$ & $92.66$ & $73.82$ & $98.32$ & $72.64$ & $100.00$ & $87.79$ \\
        \multicolumn{2}{l|}{SHOT$++$~\cite{liang2021source}} & $93.67$ & $92.45$ & $74.76$ & $98.11$ & $75.47$ & $99.00$ & $88.91$ \\
        \multicolumn{2}{l|}{AaD~\cite{yang2022attracting}} & $93.33$ & $93.08$ & $73.94$ & $97.90$ & $74.35$ & $99.67$ & $88.71$ \\
        \multicolumn{2}{l|}{CoWA-JMDS~\cite{lee2022confidence}} & $94.67$ & $94.97$ & $76.00$ & $98.53$ & $77.37$ & $99.67$ & $\underline{90.20}$ \\
        \multicolumn{2}{l|}{NRC~\cite{yang2021exploiting}} & $91.67$ & $91.82$ & $74.94$ & $98.11$ & $75.65$ & $99.33$ & $88.59$ \\
        \multicolumn{2}{l|}{G-SFDA~\cite{yang2021generalized}} & $88.67$ & $89.52$ & $73.47$ & $97.27$ & $72.99$ & $100.00$ & $86.99$ \\
        \midrule
        \multirow{3}{*}{FT} & $1$ & $85.67$ & $86.37$ & $64.83$ & $95.60$ & $65.13$ & $99.67$ & $82.88$ \\ %
        & $3$ & $92.67$ & $92.87$ & $72.64$ & $95.60$ & $72.87$ & $98.67$ & $87.55$ \\
        & $5$ & $94.67$ & $95.18$ & $77.31$ & $96.44$ & $77.83$ & $99.00$ & $\bold{90.07}$ \\
        \midrule
        \multirow{3}{*}{LP-FT} & $1$ & $84.67$ & $88.68$ & $65.90$ & $93.50$ & $65.37$ & $99.00$ & $82.85$ \\
        & $3$ & $93.00$ & $93.92$ & $76.42$ & $95.39$ & $75.18$ & $99.67$ & $\bold{88.93}$ \\
        & $5$ & $96.00$ & $96.65$ & $78.66$ & $97.69$ & $77.60$ & $99.00$ & $\textcolor{red}{\bold{90.93}}$ \\
        \bottomrule
    \end{tabular} 
    }
    \end{center}
    \caption{Accuracies of SFUDA and few-shot FT methods on the Office31 dataset. A, D, and W denote Amazon, DSLR, and Webcam, respectively. The numbers next to FT and LP-FT represent the number of labeled images per class used for training. \underline{Underline} indicates the best performance among SFUDA methods. \textbf{Bold} and \textcolor{red}{\textbf{Bold Red}} indicate fine-tuning performance surpassing the average performance and the best performance of SFUDA methods, respectively.}
    \label{tab:detail_o31}
\end{table}

\begin{table}[H]
    \begin{center}
    \resizebox{0.95\textwidth}{!}{%
    \begin{tabular}{lc|cccccccccccccc}
        \toprule
        Method & & \multicolumn{1}{c}{Ar→Cl} & \multicolumn{1}{c}{Ar→Pr} & \multicolumn{1}{c}{Ar→Rw} & \multicolumn{1}{c}{Cl→Ar} & \multicolumn{1}{c}{Cl→Pr} & \multicolumn{1}{c}{Cl→Rw} & \multicolumn{1}{c}{Pr→Ar} & \multicolumn{1}{c}{Pr→Cl} & \multicolumn{1}{c}{Pr→Rw} & \multicolumn{1}{c}{Rw→Ar} & \multicolumn{1}{c}{Rw→Cl} & \multicolumn{1}{c}{Rw→Pr} & \multicolumn{1}{c}{Avg} \\
        \midrule
        \multicolumn{2}{l|}{No adapt} & $48.19$ & $64.56$ & $73.20$ & $49.93$ & $61.64$ & $62.16$ & $49.79$ & $42.31$ & $72.21$ & $61.59$ & $48.30$ & $77.25$ & $59.26$ \\
        \multicolumn{2}{l|}{SHOT~\cite{liang2020we}} & $54.91$ & $75.45$ & $79.97$ & $64.68$ & $75.60$ & $77.14$ & $64.61$ & $53.68$ & $81.38$ & $71.74$ & $57.39$ & $82.28$ & $69.90$ \\
        \multicolumn{2}{l|}{SHOT$++$~\cite{liang2021source}} & $56.05$ & $77.78$ & $80.12$ & $65.57$ & $78.90$ & $77.37$ & $65.30$ & $53.61$ & $80.70$ & $71.33$ & $56.51$ & $83.56$ & $70.57$ \\
        \multicolumn{2}{l|}{AaD~\cite{yang2022attracting}} & $56.16$ & $77.89$ & $79.74$ & $63.72$ & $79.99$ & $78.52$ & $63.10$ & $56.13$ & $80.81$ & $69.75$ & $57.85$ & $84.83$ & $70.71$ \\
        \multicolumn{2}{l|}{CoWA-JMDS~\cite{lee2022confidence}} & $57.47$ & $78.60$ & $81.50$ & $67.76$ & $79.43$ & $79.89$ & $65.77$ & $56.74$ & $82.27$ & $71.40$ & $59.91$ & $84.08$ & $\underline{72.07}$ \\
        \multicolumn{2}{l|}{NRC~\cite{yang2021exploiting}} & $55.59$ & $76.76$ & $79.74$ & $65.50$ & $76.80$ & $77.41$ & $65.09$ & $54.94$ & $80.51$ & $71.12$ & $56.97$ & $82.51$ & $70.25$ \\
        \multicolumn{2}{l|}{G-SFDA~\cite{yang2021generalized}} & $54.98$ & $75.19$ & $79.82$ & $64.27$ & $73.65$ & $75.88$ & $63.44$ & $52.58$ & $80.58$ & $70.30$ & $57.58$ & $82.36$ & $69.22$ \\
        \midrule
        \multirow{4}{*}{FT} & $1$ & $53.38$ & $68.92$ & $74.58$ & $57.82$ & $70.76$ & $68.61$ & $57.13$ & $50.14$ & $74.81$ & $64.81$ & $54.72$ & $77.66$ & $64.45$ \\ %
        & $3$ & $58.19$ & $77.97$ & $77.45$ & $63.78$ & $78.57$ & $74.31$ & $65.09$ & $58.50$ & $77.98$ & $68.73$ & $61.59$ & $82.43$ & $70.38$ \\
        & $5$ & $62.54$ & $80.03$ & $78.86$ & $66.87$ & $80.29$ & $77.83$ & $67.28$ & $64.57$ & $80.20$ & $70.99$ & $66.63$ & $84.38$ & $\textcolor{red}{\bold{73.37}}$\\
        & $10$ & $67.81$ & $84.83$ & $81.19$ & $71.95$ & $85.70$ & $80.58$ & $72.43$ & $68.38$ & $81.80$ & $74.28$ & $69.15$ & $85.77$ & $\textcolor{red}{\bold{76.99}}$ \\
        \midrule
        \multirow{4}{*}{LP-FT} & $1$ & $53.38$ & $69.78$ & $74.77$ & $56.72$ & $69.74$ & $68.12$ & $57.55$ & $48.45$ & $73.70$ & $63.51$ & $51.24$ & $76.58$ & $63.63$ \\
        & $3$ & $59.30$ & $80.74$ & $77.22$ & $65.98$ & $79.58$ & $75.73$ & $66.94$ & $62.35$ & $78.82$ & $70.16$ & $63.50$ & $82.96$ & $\bold{71.94}$ \\
        & $5$ & $64.03$ & $82.77$ & $80.16$ & $69.27$ & $81.12$ & $77.98$ & $70.30$ & $64.41$ & $80.27$ & $72.70$ & $66.70$ & $84.87$ & $\textcolor{red}{\bold{74.55}}$ \\
        & $10$ & $67.73$ & $86.79$ & $83.95$ & $74.35$ & $86.26$ & $81.99$ & $74.08$ & $69.49$ & $83.87$ & $76.89$ & $69.95$ & $87.61$ & $\textcolor{red}{\bold{78.58}}$ \\
        \bottomrule
    \end{tabular} 
    }
    \end{center}
    \caption{Accuracies of SFUDA and few-shot FT methods on the OfficeHome dataset. Ar, Cl, Pr, and Rw denote Art, Clipart, Product, and Real World, respectively. Other notations are the same as in Table~\ref{tab:detail_o31}.}
    \label{tab:detail_oh}
\end{table}

\begin{table}[H]
    \begin{center}
    \resizebox{0.85\textwidth}{!}{%
    \begin{tabular}{lc|cccccccccccccc}
        \toprule
        Method & & \multicolumn{1}{c}{plane} & \multicolumn{1}{c}{bcycl} & \multicolumn{1}{c}{bus} & \multicolumn{1}{c}{car} & \multicolumn{1}{c}{horse} & \multicolumn{1}{c}{knife} & \multicolumn{1}{c}{mcycl} & \multicolumn{1}{c}{person} & \multicolumn{1}{c}{plant} & \multicolumn{1}{c}{sktbrd} & \multicolumn{1}{c}{train} & \multicolumn{1}{c}{truck} & \multicolumn{1}{c}{Avg}\\
        \midrule
        \multicolumn{2}{l|}{No adapt} & $60.62$ & $19.60$ & $52.24$ & $73.26$ & $64.99$ & $4.45$ & $82.75$ & $16.63$ & $73.15$ & $35.66$ & $82.46$ & $7.53$ & $47.78$ \\
        \multicolumn{2}{l|}{SHOT~\cite{liang2020we}} & $95.16$ & $89.06$ & $79.39$ & $53.66$ & $92.79$ & $94.38$ & $79.38$ & $80.83$ & $89.49$ & $86.62$ & $86.74$ & $53.72$ & $81.77$ \\
        \multicolumn{2}{l|}{SHOT$++$~\cite{liang2021source}} & $97.21$ & $86.42$ & $89.48$ & $84.93$ & $97.86$ & $97.51$ & $92.66$ & $84.00$ & $96.96$ & $92.47$ & $93.55$ & $29.04$ & $\underline{86.84}$ \\
        \multicolumn{2}{l|}{AaD~\cite{yang2022attracting}} & $96.43$ & $91.02$ & $88.95$ & $80.40$ & $96.56$ & $96.64$ & $89.71$ & $79.83$ & $94.39$ & $90.87$ & $91.01$ & $47.71$ & $86.68$ \\
        \multicolumn{2}{l|}{CoWA-JMDS~\cite{lee2022confidence}} & $95.33$ & $88.49$ & $82.12$ & $71.21$ & $95.70$ & $97.99$ & $89.79$ & $85.54$ & $92.82$ & $90.35$ & $87.29$ & $56.07$ & $86.06$ \\
        \multicolumn{2}{l|}{NRC~\cite{yang2021exploiting}} & $96.98$ & $93.19$ & $83.97$ & $60.60$ & $96.23$ & $95.50$ & $81.48$ & $79.75$ & $93.56$ & $91.96$ & $90.67$ & $59.79$ & $85.31$ \\
        \multicolumn{2}{l|}{G-SFDA~\cite{yang2021generalized}} & $96.48$ & $88.20$ & $84.40$ & $70.64$ & $95.92$ & $96.38$ & $89.04$ & $79.46$ & $93.78$ & $91.01$ & $89.22$ & $42.07$ & $84.72$ \\
        \midrule
        \multirow{4}{*}{FT} & $10$ & $95.11$ & $81.73$ & $82.91$ & $64.38$ & $94.03$ & $91.16$ & $86.37$ & $74.04$ & $89.71$ & $78.29$ & $85.48$ & $51.17$ & $81.20$ \\ %
        & $20$ & $95.47$ & $80.62$ & $82.52$ & $64.90$ & $94.39$ & $91.33$ & $90.39$ & $80.79$ & $90.62$ & $82.24$ & $87.72$ & $58.77$ & $83.31$ \\
        & $30$ & $95.20$ & $85.13$ & $81.66$ & $67.95$ & $94.56$ & $95.02$ & $87.86$ & $81.08$ & $90.51$ & $84.50$ & $88.98$ & $58.92$ & $84.28$ \\
        & $50$ & $95.93$ & $84.17$ & $78.93$ & $71.65$ & $94.53$ & $93.82$ & $90.77$ & $83.33$ & $94.10$ & $88.82$ & $90.71$ & $64.41$ & $\bold{85.93}$ \\
        \midrule
        \multirow{4}{*}{LP-FT} & $10$ & $93.74$ & $80.19$ & $79.28$ & $67.10$ & $93.07$ & $91.08$ & $87.37$ & $76.79$ & $89.34$ & $77.92$ & $83.27$ & $55.77$ & $81.24$ \\
        & $20$ & $94.79$ & $81.77$ & $79.21$ & $66.67$ & $94.28$ & $93.65$ & $90.14$ & $80.33$ & $89.78$ & $83.55$ & $89.49$ & $58.65$ & $83.53$ \\
        & $30$ & $95.38$ & $87.19$ & $80.60$ & $69.29$ & $95.34$ & $93.01$ & $87.12$ & $82.58$ & $90.29$ & $85.96$ & $90.55$ & $60.48$ & $84.82$ \\
        & $50$ & $96.02$ & $83.98$ & $80.21$ & $72.59$ & $94.56$ & $93.41$ & $90.91$ & $83.04$ & $94.25$ & $89.18$ & $90.40$ & $65.50$ & $\bold{86.17}$ \\
        \bottomrule
    \end{tabular} 
    }
    \end{center}
    \caption{Accuracies of SFUDA and few-shot FT methods on the VisDA dataset (Synthesis $\rightarrow$ Real). Other notations are the same as in Table~\ref{tab:detail_o31}.}
    \label{tab:detail_visda}
\end{table}

\begin{table}[H]
    \begin{center}
    \resizebox{0.85\textwidth}{!}{%
    \begin{tabular}{lc|cccccccccccccc}
        \toprule
        Method & & \multicolumn{1}{c}{C$\rightarrow$L} & \multicolumn{1}{c}{C$\rightarrow$S} & \multicolumn{1}{c}{C$\rightarrow$V} & \multicolumn{1}{c}{L$\rightarrow$C} & \multicolumn{1}{c}{L$\rightarrow$S} & \multicolumn{1}{c}{L$\rightarrow$V} & \multicolumn{1}{c}{S$\rightarrow$C} & \multicolumn{1}{c}{S$\rightarrow$L} & \multicolumn{1}{c}{S$\rightarrow$V} & \multicolumn{1}{c}{V$\rightarrow$C} & \multicolumn{1}{c}{V$\rightarrow$L} & \multicolumn{1}{c}{V$\rightarrow$S} & \multicolumn{1}{c}{Avg}\\
        \midrule
        \multicolumn{2}{l|}{No adapt} & $47.80$ & $53.29$ & $64.57$ & $51.26$ & $40.66$ & $55.10$ & $58.12$ & $36.26$ & $55.28$ & $97.75$ & $48.77$ & $71.96$ & $56.74$ \\
        \multicolumn{2}{l|}{SHOT~\cite{liang2020we}} & $41.55$ & $55.65$ & $75.89$ & $65.31$ & $60.82$ & $74.78$ & $87.74$ & $41.11$ & $82.73$ & $89.80$ & $45.81$ & $65.71$ & $65.58$ \\
        \multicolumn{2}{l|}{SHOT$++$~\cite{liang2021source}} & $41.72$ & $58.35$ & $75.58$ & $70.00$ & $56.89$ & $76.21$ & $71.61$ & $39.45$ & $80.72$ & $96.86$ & $43.73$ & $60.07$ & $64.27$ \\
        \multicolumn{2}{l|}{AaD~\cite{yang2022attracting}} & $37.69$ & $57.20$ & $75.47$ & $59.47$ & $48.81$ & $67.45$ & $84.03$ & $34.81$ & $72.84$ & $43.38$ & $40.19$ & $54.80$ & $56.35$ \\
        \multicolumn{2}{l|}{CoWA-JMDS~\cite{lee2022confidence}} & $46.05$ & $58.41$ & $81.09$ & $85.58$ & $64.07$ & $78.22$ & $95.91$ & $48.12$ & $82.70$ & $99.40$ & $50.80$ & $65.27$ & $\underline{71.30}$ \\
        \multicolumn{2}{l|}{NRC~\cite{yang2021exploiting}} & $39.88$ & $55.85$ & $75.45$ & $64.69$ & $54.09$ & $74.76$ & $78.42$ & $40.28$ & $82.18$ & $90.10$ & $41.73$ & $62.82$ & $63.35$ \\
        \multicolumn{2}{l|}{G-SFDA~\cite{yang2021generalized}} & $42.64$ & $54.87$ & $73.46$ & $82.25$ & $51.43$ & $72.01$ & $74.56$ & $45.84$ & $82.68$ & $88.85$ & $49.05$ & $64.27$ & $65.16$ \\
        \midrule
        \multirow{4}{*}{FT} & $1$ & $48.78$ & $49.78$ & $73.24$ & $94.39$ & $44.95$ & $54.85$ & $94.12$ & $40.18$ & $65.76$ & $98.56$ & $46.68$ & $61.32$ & $\bold{64.39}$ \\ %
        & $3$ & $49.11$ & $64.73$ & $75.01$ & $98.73$ & $56.13$ & $73.46$ & $98.69$ & $49.56$ & $72.31$ & $98.73$ & $52.28$ & $69.67$ & $\textcolor{red}{\bold{71.53}}$ \\
        & $5$ & $47.53$ & $65.84$ & $78.04$ & $99.16$ & $59.13$ & $73.67$ & $99.16$ & $52.12$ & $74.89$ & $99.30$ & $49.77$ & $66.71$ & $\textcolor{red}{\bold{72.11}}$ \\
        & $10$ & $50.53$ & $66.72$ & $79.49$ & $99.71$ & $65.61$ & $77.56$ & $99.68$ & $54.12$ & $79.46$ & $99.72$ & $51.19$ & $69.75$ & $\textcolor{red}{\bold{74.46}}$ \\
        \midrule
        \multirow{4}{*}{LP-FT} & $1$ & $38.07$ & $52.68$ & $73.07$ & $96.35$ & $47.03$ & $52.44$ & $96.72$ & $48.93$ & $65.28$ & $98.32$ & $50.49$ & $58.41$ & $\bold{64.82}$ \\
        & $3$ & $50.80$ & $66.18$ & $79.57$ & $98.88$ & $61.90$ & $73.24$ & $98.70$ & $51.32$ & $77.13$ & $99.16$ & $50.38$ & $69.24$ & $\textcolor{red}{\bold{73.04}}$ \\
        & $5$ & $44.83$ & $65.49$ & $80.25$ & $99.30$ & $63.09$ & $73.38$ & $98.88$ & $51.43$ & $78.15$ & $99.68$ & $50.79$ & $71.63$ & $\textcolor{red}{\bold{73.07}}$ \\
        & $10$ & $50.99$ & $66.72$ & $81.06$ & $99.72$ & $62.24$ & $77.99$ & $99.82$ & $55.46$ & $80.50$ & $99.72$ & $53.03$ & $67.00$ & $\textcolor{red}{\bold{74.52}}$ \\
        \bottomrule
    \end{tabular} 
    }
    \end{center}
    \caption{Per-class accuracies of SFUDA and few-shot FT methods on the VLCS dataset. C, L, P, and V denote Caltech101, LabelMe, Sun09, and VOC2007, respectively. Other notations are the same as in Table~\ref{tab:detail_o31}.}
    \label{tab:detail_vlcs}
\end{table}

\begin{table}[H]
    \begin{center}
    \resizebox{1\textwidth}{!}{%
    \begin{tabular}{lc|cccccccccccccc}
        \toprule
        Method & & \multicolumn{1}{c}{L100$\rightarrow$L38} & \multicolumn{1}{c}{L100$\rightarrow$L43} & \multicolumn{1}{c}{L100$\rightarrow$L46} & \multicolumn{1}{c}{L38$\rightarrow$L100} & \multicolumn{1}{c}{L38$\rightarrow$L43} & \multicolumn{1}{c}{L38$\rightarrow$L46} & \multicolumn{1}{c}{L43$\rightarrow$L100} & \multicolumn{1}{c}{L43$\rightarrow$L38} & \multicolumn{1}{c}{L43$\rightarrow$L46} & \multicolumn{1}{c}{L46$\rightarrow$L100} & \multicolumn{1}{c}{L46$\rightarrow$L38} & \multicolumn{1}{c}{L46$\rightarrow$L43} & \multicolumn{1}{c}{Avg}\\
        \midrule
        \multicolumn{2}{l|}{No adapt} & $26.24$ & $20.34$ & $27.08$ & $29.31$ & $31.35$ & $31.59$ & $24.11$ & $44.09$ & $38.66$ & $33.60$ & $21.57$ & $22.18$ & $29.18$ \\
        \multicolumn{2}{l|}{SHOT~\cite{liang2020we}} & $20.07$ & $23.82$ & $28.48$ & $35.95$ & $28.98$ & $13.57$ & $26.20$ & $14.52$ & $32.72$ & $34.32$ & $12.63$ & $37.41$ & $25.72$ \\
        \multicolumn{2}{l|}{SHOT$++$~\cite{liang2021source}} & $29.32$ & $22.08$ & $25.47$ & $22.76$ & $31.79$ & $18.44$ & $33.31$ & $22.61$ & $25.60$ & $35.82$ & $13.00$ & $44.59$ & $27.07$ \\
        \multicolumn{2}{l|}{AaD~\cite{yang2022attracting}} & $17.15$ & $17.35$ & $22.14$ & $24.56$ & $28.10$ & $13.30$ & $28.92$ & $23.34$ & $23.06$ & $31.56$ & $7.35$ & $34.59$ & $22.62$ \\
        \multicolumn{2}{l|}{CoWA-JMDS~\cite{lee2022confidence}} & $33.08$ & $31.42$ & $26.35$ & $36.26$ & $38.33$ & $19.25$ & $28.23$ & $13.57$ & $26.62$ & $32.52$ & $9.96$ & $47.56$ & $28.60$ \\
        \multicolumn{2}{l|}{NRC~\cite{yang2021exploiting}} & $19.28$ & $22.70$ & $29.47$ & $38.46$ & $26.88$ & $14.94$ & $30.78$ & $22.64$ & $32.16$ & $28.86$ & $11.04$ & $39.04$ & $26.35$ \\
        \multicolumn{2}{l|}{G-SFDA~\cite{yang2021generalized}} & $21.59$ & $29.14$ & $38.17$ & $38.38$ & $27.03$ & $22.40$ & $40.86$ & $17.42$ & $33.29$ & $35.01$ & $16.25$ & $52.55$ & $\underline{31.01}$ \\
        \midrule
        \multirow{3}{*}{FT} & $1$ & $32.07$ & $30.27$ & $27.48$ & $44.81$ & $28.52$ & $29.05$ & $48.33$ & $48.79$ & $34.15$ & $55.72$ & $46.84$ & $44.06$ & $\textcolor{red}{\bold{39.17}}$ \\ %
        & $3$ & $46.06$ & $47.63$ & $32.02$ & $58.63$ & $41.67$ & $34.91$ & $60.22$ & $52.58$ & $42.39$ & $62.88$ & $54.79$ & $47.56$ & $\textcolor{red}{\bold{48.44}}$ \\
        & $5$ & $52.42$ & $41.66$ & $50.02$ & $63.81$ & $38.57$ & $47.84$ & $66.15$ & $56.70$ & $51.43$ & $68.89$ & $56.67$ & $61.20$ & $\textcolor{red}{\bold{54.61}}$ \\
        \midrule
        \multirow{3}{*}{LP-FT} & $1$ & $36.03$ & $35.07$ & $34.65$ & $47.83$ & $31.85$ & $35.72$ & $47.51$ & $43.97$ & $35.66$ & $54.63$ & $43.63$ & $44.92$ & $\textcolor{red}{\bold{40.96}}$ \\
        & $3$ & $47.50$ & $38.51$ & $38.12$ & $58.06$ & $39.76$ & $41.70$ & $60.59$ & $49.19$ & $45.00$ & $62.87$ & $54.28$ & $52.13$ & $\textcolor{red}{\bold{48.98}}$ \\
        & $5$ & $54.28$ & $47.45$ & $46.93$ & $63.59$ & $41.32$ & $48.97$ & $64.24$ & $55.87$ & $54.37$ & $68.36$ & $55.69$ & $63.77$ & $\textcolor{red}{\bold{55.40}}$ \\
        \bottomrule
    \end{tabular} 
    }
    \end{center}
    \caption{Per-class accuracies of SFUDA and few-shot FT methods on the Terra dataset. L100, L38, L43, and L46 denote different locations (i.e., domains). Other notations are the same as in Table~\ref{tab:detail_o31}.}
    \label{tab:detail_terra}
\end{table}

\begin{table}[H]
    \begin{center}
    \resizebox{0.55\textwidth}{!}{%
    \begin{tabular}{lc|cccccccc}
        \toprule
        Method & & \multicolumn{1}{c}{A→D} & \multicolumn{1}{c}{A→W} & \multicolumn{1}{c}{D→A} & \multicolumn{1}{c}{D→W} & \multicolumn{1}{c}{W→A} & \multicolumn{1}{c}{W→D} & \multicolumn{1}{c}{Avg}\\
        \midrule
        \multicolumn{2}{l|}{SHOT~\cite{liang2020we}} & $83.33$ & $86.06$ & $76.76$ & $100.00$ & $81.08$ & $100.00$ & $87.87$ \\
        \multicolumn{2}{l|}{SHOT$++$~\cite{liang2021source}} & $90.20$ & $81.21$ & $80.54$ & $99.39$ & $82.52$ & $98.04$ & $88.65$ \\
        \multicolumn{2}{l|}{AaD~\cite{yang2022attracting}} & $88.24$ & $87.27$ & $75.68$ & $100.00$ & $81.80$ & $99.02$ & $88.67$ \\
        \multicolumn{2}{l|}{CoWA-JMDS~\cite{lee2022confidence}} & $89.22$ & $87.27$ & $78.20$ & $100.00$ & $81.98$ & $100.00$ & $89.45$ \\
        \multicolumn{2}{l|}{NRC~\cite{yang2021exploiting}} & $90.20$ & $85.46$ & $75.32$ & $97.58$ & $81.80$ & $99.02$ & $88.23$ \\
        \multicolumn{2}{l|}{G-SFDA~\cite{yang2021generalized}} & $87.77$ & $89.16$ & $79.55$ & $99.17$ & $84.17$ & $100.00$ & $\underline{89.97}$ \\
        \midrule
        \multirow{2}{*}{FT} & $1$ & $95.10$ & $89.09$ & $81.80$ & $95.76$ & $82.52$ & $97.06$ & $\textcolor{red}{\bold{90.22}}$ \\ %
        & $3$ & $95.10$ & $93.33$ & $88.29$ & $96.97$ & $87.75$ & $98.04$ & $\textcolor{red}{\bold{93.25}}$ \\
        \midrule
        \multirow{2}{*}{LP-FT} & $1$ & $95.10$ & $89.09$ & $83.78$ & $95.76$ & $84.15$ & $98.04$ & $\textcolor{red}{\bold{90.99}}$ \\
        & $3$ & $96.08$ & $95.15$ & $87.21$ & $96.97$ & $89.55$ & $97.06$ & $\textcolor{red}{\bold{93.67}}$ \\
        \bottomrule
    \end{tabular} 
    }
    \end{center}
    \caption{Accuracies of SFUDA and few-shot FT methods on Office31 under OoD scenarios. Notations are the same as in Table~\ref{tab:detail_o31}.}
    \label{tab:ood_o31}
\end{table}

\begin{table}[H]
    \begin{center}
    \resizebox{1\textwidth}{!}{%
    \begin{tabular}{lc|cccccccccccccc}
        \toprule
        Method & & \multicolumn{1}{c}{Ar→Cl} & \multicolumn{1}{c}{Ar→Pr} & \multicolumn{1}{c}{Ar→Rw} & \multicolumn{1}{c}{Cl→Ar} & \multicolumn{1}{c}{Cl→Pr} & \multicolumn{1}{c}{Cl→Rw} & \multicolumn{1}{c}{Pr→Ar} & \multicolumn{1}{c}{Pr→Cl} & \multicolumn{1}{c}{Pr→Rw} & \multicolumn{1}{c}{Rw→Ar} & \multicolumn{1}{c}{Rw→Cl} & \multicolumn{1}{c}{Rw→Pr} & \multicolumn{1}{c}{Avg} \\
        \midrule
        \multicolumn{2}{l|}{SHOT~\cite{liang2020we}} & $51.76$ & $70.68$ & $74.17$ & $65.23$ & $68.78$ & $73.63$ & $63.58$ & $51.57$ & $75.35$ & $72.84$ & $54.89$ & $77.41$ & $66.66$ \\
        \multicolumn{2}{l|}{SHOT$++$~\cite{liang2021source}} & $52.99$ & $71.89$ & $73.10$ & $62.76$ & $69.78$ & $72.78$ & $62.96$ & $50.05$ & $74.60$ & $72.22$ & $52.23$ & $80.32$ & $66.31$ \\
        \multicolumn{2}{l|}{AaD~\cite{yang2022attracting}} & $54.32$ & $69.58$ & $69.88$ & $63.99$ & $64.26$ & $64.85$ & $61.73$ & $49.48$ & $73.20$ & $69.55$ & $52.13$ & $76.31$ & $64.11$ \\
        \multicolumn{2}{l|}{CoWA-JMDS~\cite{lee2022confidence}} & $53.56$ & $72.99$ & $73.85$ & $66.25$ & $68.97$ & $73.10$ & $65.02$ & $51.95$ & $74.28$ & $71.40$ & $55.65$ & $80.62$ & $67.30$ \\
        \multicolumn{2}{l|}{NRC~\cite{yang2021exploiting}} & $51.47$ & $72.09$ & $77.28$ & $69.75$ & $70.28$ & $75.78$ & $67.08$ & $52.52$ & $77.71$ & $73.04$ & $54.89$ & $79.92$ & $\underline{68.48}$ \\
        \multicolumn{2}{l|}{G-SFDA~\cite{yang2021generalized}} & $53.52$ & $71.28$ & $74.06$ & $60.10$ & $69.66$ & $71.62$ & $58.51$ & $53.83$ & $73.50$ & $72.45$ & $60.48$ & $81.16$ & $66.68$ \\
        \midrule
        \multirow{2}{*}{FT} & $1$ & $52.90$ & $73.59$ & $76.21$ & $57.20$ & $73.70$ & $73.42$ & $60.08$ & $56.69$ & $75.46$ & $72.22$ & $58.02$ & $79.92$ & $\bold{67.45}$ \\ %
        & $3$ & $62.87$ & $83.43$ & $80.49$ & $68.72$ & $82.53$ & $78.88$ & $68.11$ & $64.86$ & $78.35$ & $76.75$ & $66.76$ & $85.94$ & $\textcolor{red}{\bold{74.81}}$ \\
        \midrule
        \multirow{2}{*}{LP-FT} & $1$ & $51.76$ & $75.90$ & $74.92$ & $59.47$ & $75.81$ & $72.34$ & $56.58$ & $53.28$ & $76.10$ & $72.43$ & $56.31$ & $80.42$ & $\bold{67.11}$ \\
        & $3$ & $64.39$ & $84.94$ & $81.78$ & $71.61$ & $83.64$ & $79.85$ & $70.37$ & $63.82$ & $80.28$ & $75.10$ & $66.38$ & $85.34$ & $\textcolor{red}{\bold{75.63}}$ \\
        \bottomrule
    \end{tabular} 
    }
    \end{center}
    \caption{Accuracies of SFUDA and few-shot FT methods on OfficeHome under OoD scenarios. Ar, Cl, Pr, and Rw denote Art, Clipart, Product, and Real World, respectively. Other notations are the same as in Table~\ref{tab:detail_o31}.}
    \label{tab:ood_oh}
\end{table}

\begin{table}[H]
    \begin{center}
    \resizebox{0.55\textwidth}{!}{%
    \begin{tabular}{lc|cccccccccccccc}
        \toprule
        Method & & \multicolumn{1}{c}{bcycl} & \multicolumn{1}{c}{bus} & \multicolumn{1}{c}{horse} & \multicolumn{1}{c}{knife} & \multicolumn{1}{c}{train} & \multicolumn{1}{c}{truck} & \multicolumn{1}{c}{Avg}\\
        \midrule
        \multicolumn{2}{l|}{SHOT~\cite{liang2020we}} & $89.40$ & $89.59$ & $95.27$ & $98.07$ & $86.90$ & $43.75$ & $83.83$ \\
        \multicolumn{2}{l|}{SHOT$++$~\cite{liang2021source}} & $94.92$ & $87.74$ & $96.13$ & $98.80$ & $92.05$ & $32.10$ & $83.62$ \\
        \multicolumn{2}{l|}{AaD~\cite{yang2022attracting}} & $94.48$ & $85.32$ & $96.09$ & $98.55$ & $87.09$ & $67.42$ & $\underline{88.16}$ \\
        \multicolumn{2}{l|}{CoWA-JMDS~\cite{lee2022confidence}} & $87.05$ & $92.32$ & $95.17$ & $99.52$ & $70.41$ & $32.25$ & $79.45$ \\
        \multicolumn{2}{l|}{NRC~\cite{yang2021exploiting}} & $94.34$ & $89.84$ & $96.84$ & $97.11$ & $92.64$ & $44.42$ & $85.86$ \\
        \multicolumn{2}{l|}{G-SFDA~\cite{yang2021generalized}} & $91.08$ & $88.77$ & $95.74$ & $98.31$ & $90.16$ & $0.00$ & $77.34$ \\
        \midrule
        \multirow{2}{*}{FT} & $3$ & $92.57$ & $72.32$ & $91.12$ & $97.75$ & $88.55$ & $63.57$ & $\bold{84.31}$ \\
        & $10$ & $90.60$ & $75.69$ & $94.17$ & $99.12$ & $84.45$ & $71.56$ & $\bold{85.93}$ \\
        \midrule
        \multirow{2}{*}{LP-FT} & $3$ & $91.85$ & $72.49$ & $92.18$ & $97.83$ & $86.15$ & $57.12$ & $82.93$ \\
        & $10$ & $92.23$ & $73.38$ & $95.02$ & $98.63$ & $84.30$ & $76.61$ & $\bold{86.70}$ \\
        \bottomrule
    \end{tabular} 
    }
    \end{center}
    \caption{Accuracies of SFUDA and few-shot FT methods on VisDA (Synthesis $\rightarrow$ Real) under OoD scenarios. Notations are the same as in Table~\ref{tab:detail_o31}.}
    \label{tab:ood_visda}
\end{table}

\begin{table}[H]
    \begin{center}
    \resizebox{0.65\textwidth}{!}{%
    \begin{tabular}{lc|cccccccccccccc}
        \toprule
        Method & & \multicolumn{1}{c}{Cl→Pr} & \multicolumn{1}{c}{Cl→Rw} & \multicolumn{1}{c}{Pr→Cl} & \multicolumn{1}{c}{Pr→Rw} & \multicolumn{1}{c}{Rw→Cl} & \multicolumn{1}{c}{Rw→Pr} & \multicolumn{1}{c}{Avg} \\
        \midrule
        \multicolumn{2}{l|}{SHOT~\cite{liang2020we}} & $61.44$ & $61.73$ & $47.98$ & $71.60$ & $51.35$ & $75.91$ & $61.67$ \\
        \multicolumn{2}{l|}{SHOT$++$~\cite{liang2021source}} & $62.15$ & $59.59$ & $44.12$ & $69.42$ & $46.35$ & $75.30$ & $59.49$ \\
        \multicolumn{2}{l|}{AaD~\cite{yang2022attracting}} & $64.60$ & $64.72$ & $46.61$ & $72.53$ & $50.87$ & $74.09$ & $62.24$ \\
        \multicolumn{2}{l|}{CoWA-JMDS~\cite{lee2022confidence}} & $66.57$ & $64.02$ & $46.99$ & $75.40$ & $49.99$ & $75.84$ & $\underline{63.14}$ \\
        \multicolumn{2}{l|}{NRC~\cite{yang2021exploiting}} & $60.49$ & $60.22$ & $42.78$ & $67.40$ & $45.75$ & $71.25$ & $57.98$ \\
        \multicolumn{2}{l|}{G-SFDA~\cite{yang2021generalized}} & $58.09$ & $56.35$ & $41.66$ & $69.48$ & $45.26$ & $71.47$ & $57.05$ \\
        \midrule
        \multirow{2}{*}{FT} & $1$ & $65.76$ & $62.71$ & $48.15$ & $70.68$ & $52.99$ & $69.50$ & $\bold{61.63}$ \\ %
        & $3$ & $74.92$ & $69.71$ & $57.20$ & $74.11$ & $60.66$ & $77.27$ & $\textcolor{red}{\bold{68.98}}$ \\
        \midrule
        \multirow{2}{*}{LP-FT} & $1$ & $66.92$ & $65.82$ & $48.94$ & $71.77$ & $53.84$ & $72.95$ & $\textcolor{red}{\bold{63.37}}$ \\
        & $3$ & $77.99$ & $74.04$ & $60.40$ & $75.58$ & $64.23$ & $79.89$ & $\textcolor{red}{\bold{72.02}}$ \\
        \bottomrule
    \end{tabular} 
    }
    \end{center}
    \caption{Per-class accuracies of SFUDA and few-shot FT methods on OfficeHome (RSUT). Cl, Pr, and Rw denote Clipart, Product, and Real World, respectively. Art domain is excluded following \cite{tan2020class}. Other notations are the same as in Table~\ref{tab:detail_o31}.}
    \label{tab:oh_rsut}
\end{table}

\begin{table}[H]
    \begin{center}
    \resizebox{0.9\textwidth}{!}{%
    \begin{tabular}{lc|cccccccccccccc}
        \toprule
        Method & & \multicolumn{1}{c}{plane} & \multicolumn{1}{c}{bcycl} & \multicolumn{1}{c}{bus} & \multicolumn{1}{c}{car} & \multicolumn{1}{c}{horse} & \multicolumn{1}{c}{knife} & \multicolumn{1}{c}{mcycl} & \multicolumn{1}{c}{person} & \multicolumn{1}{c}{plant} & \multicolumn{1}{c}{sktbrd} & \multicolumn{1}{c}{train} & \multicolumn{1}{c}{truck} & \multicolumn{1}{c}{Avg}\\
        \midrule
        \multicolumn{2}{l|}{SHOT~\cite{liang2020we}} & $96.30$ & $66.67$ & $73.01$ & $44.79$ & $93.88$ & $82.22$ & $85.96$ & $68.21$ & $73.13$ & $62.00$ & $58.54$ & $17.59$ & $68.52$ \\
        \multicolumn{2}{l|}{SHOT$++$~\cite{liang2021source}} & $37.03$ & $7.14$ & $44.44$ & $79.16$ & $91.16$ & $62.67$ & $92.40$ & $69.17$ & $93.79$ & $95.17$ & $88.63$ & $25.14$ & $65.49$ \\
        \multicolumn{2}{l|}{AaD~\cite{yang2022attracting}} & $70.37$ & $0.00$ & $20.63$ & $66.66$ & $92.52$ & $92.89$ & $91.81$ & $78.03$ & $93.67$ & $94.75$ & $93.79$ & $42.65$ & $\underline{69.82}$ \\
        \multicolumn{2}{l|}{CoWA-JMDS~\cite{lee2022confidence}} & $21.62$ & $7.74$ & $73.72$ & $57.18$ & $92.38$ & $98.22$ & $91.06$ & $66.91$ & $82.43$ & $62.68$ & $51.19$ & $27.22$ & $61.03$ \\
        \multicolumn{2}{l|}{NRC~\cite{yang2021exploiting}} & $92.59$ & $11.91$ & $68.26$ & $43.75$ & $90.48$ & $95.55$ & $92.11$ & $67.82$ & $89.73$ & $79.92$ & $84.68$ & $13.61$ & $69.20$ \\
        \multicolumn{2}{l|}{G-SFDA~\cite{yang2021generalized}} & $81.48$ & $0.00$ & $33.33$ & $53.12$ & $93.20$ & $0.00$ & $91.81$ & $0.00$ & $97.09$ & $0.00$ & $93.35$ & $0.00$ & $45.28$ \\
        \midrule
        \multirow{2}{*}{FT} & $3$ & $100.00$ & $73.81$ & $88.89$ & $61.46$ & $89.12$ & $79.56$ & $86.55$ & $67.63$ & $74.78$ & $63.50$ & $75.56$ & $21.42$ & $\textcolor{red}{\bold{73.52}}$ \\ %
        & $10$ & $100.00$ & $80.95$ & $90.48$ & $67.71$ & $90.48$ & $93.78$ & $80.99$ & $74.76$ & $81.24$ & $84.83$ & $83.31$ & $39.11$ & $\textcolor{red}{\bold{80.64}}$ \\
        \midrule
        \multirow{2}{*}{LP-FT} & $3$ & $96.30$ & $76.19$ & $87.3$ & $65.62$ & $89.80$ & $86.67$ & $83.33$ & $63.97$ & $79.97$ & $73.58$ & $80.67$ & $23.33$ & $\textcolor{red}{\bold{75.56}}$ \\
        & $10$ & $100.00$ & $83.33$ & $92.06$ & $69.79$ & $92.52$ & $90.67$ & $85.67$ & $76.11$ & $82.00$ & $80.58$ & $85.12$ & $44.13$ & $\textcolor{red}{\bold{81.83}}$ \\
        \bottomrule
    \end{tabular} 
    }
    \end{center}
    \caption{Accuracies of SFUDA and few-shot FT methods on VisDA (RSUT). Notations are the same as in Table~\ref{tab:detail_o31}.}
    \label{tab:visda_rsut}
\end{table}

\end{document}